\documentclass{article}

    \usepackage[final]{neurips_2021}

\usepackage[utf8]{inputenc} % allow utf-8 input
\usepackage[T1]{fontenc}    % use 8-bit T1 fonts
\usepackage{hyperref}       % hyperlinks
\usepackage{url}            % simple URL typesetting
\usepackage{booktabs}       % professional-quality tables
\usepackage{amsfonts}       % blackboard math symbols
\usepackage{nicefrac}       % compact symbols for 1/2, etc.
\usepackage{microtype}      % microtypography
\usepackage{amsmath,amsfonts,bm}

\def\eqref#1{equation~\ref{#1}}
\def\1{\bm{1}}

\def\vtheta{{\bm{\theta}}}

\def\vu{{\bm{u}}}
\def\vv{{\bm{v}}}

\def\vx{{\bm{x}}}

\def\vz{{\bm{z}}}

\DeclareMathAlphabet{\mathsfit}{\encodingdefault}{\sfdefault}{m}{sl}
\SetMathAlphabet{\mathsfit}{bold}{\encodingdefault}{\sfdefault}{bx}{n}

\def\gV{{\mathcal{V}}}

\def\gZ{{\mathcal{Z}}}

\DeclareMathOperator*{\argmax}{arg\,max}

\usepackage{amsmath}
\usepackage{amssymb}
\usepackage{mathtools}
\usepackage{amsfonts}
\usepackage{xfrac}
\usepackage{float}
\usepackage{wrapfig}
\usepackage{xcolor}
\usepackage{lipsum}
\usepackage{colortbl}
\usepackage{caption}
\usepackage{multirow}
\usepackage{algorithm}
\usepackage{algorithmic}
\usepackage{subcaption}
\usepackage{dsfont}
\usepackage{hyperref}
\usepackage{cleveref}

\newcommand{\vphi}{\boldsymbol{\phi}}

\newcommand{\cS}{\mathcal{S}}

\newcommand{\bR}{\mathbb{R}}

\newcommand{\RR}{\mathds{R}}

\newcommand{\Eb}{\mathbb{E}}

\newcommand*\rmd{\mathop{}\!\mathrm{d}}

\usepackage{listings,newtxtt}

\definecolor{deepblue}{rgb}{0,0,0.6}
\definecolor{deepred}{rgb}{0.6,0,0}
\definecolor{deepgreen}{rgb}{0,0.5,0}
\lstdefinestyle{lststyle}{
language=Python,
basicstyle=\ttfamily\small,
commentstyle=\color{deepred},
otherkeywords={self},             % Add keywords here
keywordstyle=\color{deepgreen},
emph={log_add_exp, log_sum_exp, index_to_log_onehot, log_onehot_to_index, log_1_min_a, q_pred_one_timestep, q_pred, q_posterior, p_pred, categorical_kl, compute_Lt},          % Custom highlighting
emphstyle=\color{deepblue},    % Custom highlighting style
stringstyle=\color{deepred},
frame=tb,                         % Any extra options here
showstringspaces=false            % 
}
\usepackage{xcolor}
\definecolor{darkgreen}{rgb}{0.0,0.6,0.0}

\title{Argmax Flows and Multinomial Diffusion: \\Learning Categorical Distributions}

\author{%
  Emiel Hoogeboom$^1$\thanks{Equal contribution.}\,\;, $\,$Didrik Nielsen$^{2*}$, $\,$Priyank Jaini$^1$, $\,$Patrick Forr\'{e}$^3$, $\,$Max Welling$^1$ \\
  UvA-Bosch Delta Lab, University of Amsterdam$^1$, \\$\,$Technical University of Denmark$^2$, $\,$University of Amsterdam$^3$ \\
  \texttt{didni@dtu.dk, e.hoogeboom@uva.nl, p.jaini@uva.nl, } \\
  \texttt{p.d.forre@uva.nl, m.welling@uva.nl}
}
\begin{document}

\maketitle

\begin{abstract}
Generative flows and diffusion models have been predominantly trained on ordinal data, for example natural images. This paper introduces two extensions of flows and diffusion for \textit{categorical} data such as language or image segmentation: \textit{Argmax Flows} and \textit{Multinomial Diffusion}. Argmax Flows are defined by a composition of a continuous distribution (such as a normalizing flow), and an argmax function. 
To optimize this model, we learn a probabilistic inverse for the argmax that lifts the categorical data to a continuous space.
Multinomial Diffusion gradually adds categorical noise in a diffusion process, for which the generative denoising process is learned. We demonstrate that our method outperforms existing dequantization approaches on text modelling and modelling on image segmentation maps in log-likelihood. 
\end{abstract}

\section{Introduction}
% Autoregressive models
\begin{wrapfigure}{r}{.5\textwidth}
\vspace{-.5cm}
\centering
\begin{subfigure}{.499\textwidth}
\includegraphics[width=.99\linewidth]{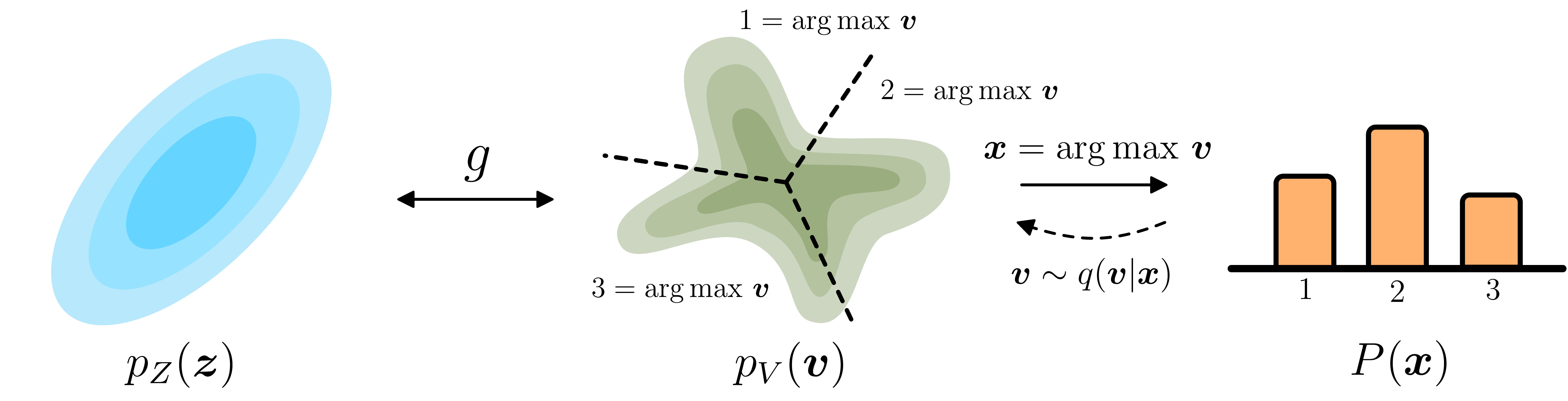}
\caption{Argmax Flow: Composition of a flow $p(\vv)$ and argmax transformation which gives the model $P(\vx)$. The flow maps from a base distribution $p(\vz)$ using a bijection $g$.}
\label{fig:overview_argmax}
\end{subfigure}
\begin{subfigure}{.499\textwidth}
\includegraphics[width=.99\textwidth]{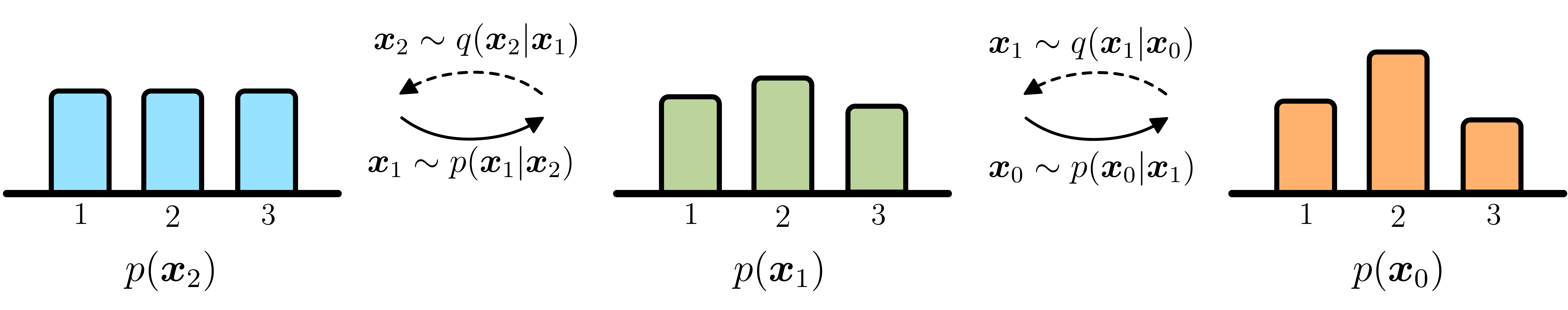}
\caption{Multinomial Diffusion: Each step $p(\vx_{t-1} | \vx_{t})$ denoises the signal starting from a uniform categorical base distribution which gives the model $p(\vx_0)$.}
\end{subfigure}
\vspace{-.25cm}
\caption{Overview of generative models.}
\label{fig:overviews}
\vspace{-.5cm}
\end{wrapfigure}
Many sources of high-dimensional data are categorical, for example language and image segmentation. Although natural images have been studied to a large extent with generative flows and diffusion models, categorical data has not had the same extensive treatment. Currently they are primarily modelled by autoregressive models, which are expensive to sample from \citep{Cooijmans2017recurrentbatch, dai2019transformerxl}.

% Problems with NFs for discrete data
Normalizing flows are attractive because they can be designed to be fast both in the evaluation and sampling direction. Typically, normalizing flows model continuous distributions. As a result, directly optimizing a flow on discrete data may lead to arbitrarily high likelihoods. In literature this problem is resolved for ordinal data by adding noise in a unit interval around the discrete value \citep{uria2013rnade,theis2016note,ho2019flow++}. However, because these methods have been designed for ordinal data, they do not work well on categorical data.

Other attractive generative models are diffusion models \citep{sohl2015diffusion}, which are fast to train due to an objective that decomposes over time steps \citep{ho2020denoising}. Diffusion models typically have a fixed diffusion process that gradually adds noise. This process is complemented by a learnable generative process that denoises the signal. \citet{song2020denoisingimplicit,nichol2021improved} have shown that diffusion models can also be designed for fast sampling. Thus far, diffusion models have been primarily trained to learn ordinal data distributions, such as natural images.

\begin{table*}[t]
\setlength{\tabcolsep}{10pt} % Default value: 6pt
\renewcommand{\arraystretch}{1.0} % Default value: 1
\caption{Surjective flow layers for applying continuous flow models to discrete data. The layers are deterministic in the generative direction, but stochastic in the inference direction.
Rounding corresponds to the commonly-used dequantization for ordinal data.}
\centering
\scalebox{.9}{
\begin{tabular}{lccl}
\toprule
\textbf{Layer} & \textbf{Generation} & \textbf{Inference} & \textbf{Applications} \\ \midrule
\multirow{2}{*}{Rounding} & \multirow{2}{*}{$\vx = \lfloor \vv \rfloor$} & $\vv \sim q(\vv|\vx)$ with support & Ordinal Data \\
 & & $\cS(\vx) = \{\vv | \vx = \lfloor \vv \rfloor \}$ & e.g. images, audio \\
\hline
\multirow{2}{*}{Argmax} & \multirow{2}{*}{$\vx = \argmax \vv$} & $\vv \sim q(\vv|\vx)$ with support & Categorical Data \\ 
 & & $\cS(\vx) = \{\vv | \vx = \argmax \vv \}$ & e.g. text, segmentation \\ 
\bottomrule
\end{tabular}}
\label{tab:surjective}
\end{table*}

Therefore, in this paper we introduce extensions of flows and diffusion models for categorical variables (depicted in Figure~\ref{fig:overviews}): \textit{i)} Argmax Flows bridge the gap between categorical data and continuous normalizing flows using an argmax transformation and a corresponding family of probabilistic inverses for the argmax. In addition \textit{ii)} we introduce Multinomial Diffusion, which is a diffusion model directly defined on categorical variables. Opposed to normalizing flows, defining diffusion for discrete variables directly does not require gradient approximations, because the diffusion trajectory is fixed. As a result of our work, generative normalizing flows and diffusion models can directly learn categorical data. 

\section{Background}
\label{sec:prelim}

\paragraph{Normalizing Flows}
Given $\gV = \mathbb{R}^d$ and $\gZ = \mathbb{R}^d$ with densities $p_{V}$ and $p_{Z}$ respectively, normalizing flows \citep{rezende2015norm} learn a bijective and differentiable transformation $g : \gZ \to \gV$ such that the change-of-variables formula gives the density at any point $\vv \in \gV$:
\small
\begin{equation}
\label{eq:cov}
    p_{V}(\vv) = p_{Z}(\vz)\cdot \left|\det \frac{\mathrm{d}\vz}{\mathrm{d}\vv} \right|,  \qquad \vv = g(\vz),
\end{equation}
\normalsize
where $p_{Z}$ can be any density (usually chosen as a standard Gaussian). Thus, normalizing flows provide a powerful framework to learn \emph{exact} density functions. However, \Cref{eq:cov} is restricted to continuous densities. 

To learn densities on ordinal discrete data (such as natural images), typically dequantization noise is added \citep{uria2013rnade, theis2016note, ho2019flow++}. \citet{nielsen2007survae} reinterpreted dequantization as a surjective flow layer $\vv \mapsto \vx$ that is deterministic in one direction ($\vx = \mathsf{round}(\vv)$) and stochastic in the other ($\vv = \vx + \vu$ where $\vu \sim q(\vu|\vx)$). Using this interpretation, dequantization can be seen as a probabilistic right-inverse for the rounding operation in the latent variable model given by:
\begin{equation*}
\small
{P(\vx) =\! \int \! P(\vx|\vv) p(\vv)\rmd\vv, \quad P(\vx|\vv) \!=\! \delta\big(\vx \!=\! \mathsf{round}(\vv)\big)},
\end{equation*}
where $\mathsf{round}$ is applied elementwise. In this case, the density model $p(\vv)$ is modeled using a normalizing flow. Learning proceeds by introducing the variational distribution $q(\vv|\vx)$ that models the probabilistic right-inverse for the rounding surjection and optimizing the evidence lower bound (ELBO):
\small
\begin{align}
\log P(\vx) &\geq \Eb_{\vv \sim q(\vv|\vx)} \left[ \log P(\vx | \vv) + \log p(\vv) - \log q(\vv|\vx) \right] = \Eb_{\vv \sim q(\vv|\vx)} \left[ \log p(\vv) - \log q(\vv|\vx) \right].
\label{eq:elbo_objective}
\end{align}
\normalsize
The last equality holds under the constraint that the support of $q(\vv|\vx)$ is enforced to be only over the region $\mathcal{S} = \{ \vv \in \RR^d : \vx = \mathsf{round}(\vv) \}$ which ensures that $P(\vx|\vv) = 1$.

\paragraph{Diffusion Models}
Given data $\vx_0$, a diffusion model \citep{sohl2015diffusion} consists of predefined variational distributions $q(\vx_t | \vx_{t-1})$ that gradually add noise over time steps $t \in \{1, \ldots, T\}$. The diffusion trajectory is defined such that $q(\vx_{t} | \vx_{t-1})$ adds a small amount of noise around $\vx_{t-1}$. This way, information is gradually destroyed such that at the final time step, $\vx_T$ carries almost no information about $\vx_0$. Their generative counterparts consists of learnable distributions $p(\vx_{t-1} | \vx_{t})$ that learn to denoise the data. 
% \small
% \begin{equation}
%     q(\vx_1, \ldots, \vx_T | \vx_0) = \prod_{t=1}^T q(\vx_{t} | \vx_{t-1}).
% \end{equation} \normalsize
%  There are generally only little or even no leanable parameters in $q$. The generative counterpart to the diffusion process is described by learnable distributions $p(\vx_{t-1} | \vx_t)$ that define the reverse (or \textit{denoising}) trajectory:
% \small\begin{equation}
% p(\vx_T, \ldots, \vx_0) = p(\vx_T) \prod_{t=1}^T p(\vx_{t-1} | \vx_t),
% \end{equation}\normalsize
When the diffusion process adds sufficiently small amounts of noise, it suffices to define the denoising trajectory using distributions that are factorized (without correlation) over the dimension axis. The distribution $p(\vx_T)$ is chosen to be similar to the distribution that the diffusion trajectory approaches. Diffusion models can be optimized using variational inference:
\begin{equation*}
    \log P(\vx_0) \geq \mathbb{E}_{x_1, \ldots x_T \sim q} 
    \Big{[} \log p(\vx_T) + \sum_{t=1}^T \log \frac{p(\vx_{t-1} | \vx_t)}{q(\vx_{t} | \vx_{t-1})} \Big{]}.
\end{equation*}

An important insight in diffusion is that by conditioning on $\vx_0$, the posterior probability $q(\vx_{t-1} | \vx_{t}, \vx_0) = q(\vx_{t} | \vx_{t-1})q(\vx_{t-1} | \vx_0) / q(\vx_t | \vx_0)$ is tractable and straightforward to compute, permitting a reformulation in terms of KL divergences that has lower variance \citep{sohl2015diffusion}. Note that $\mathrm{KL} \big{(} q(\vx_T | \vx_0) | p(\vx_T)\big{)} \approx 0$ if the diffusion trajectory $q$ is defined well:
\small
\begin{align}
\begin{split}
    &\log P(\vx_0) \geq \mathbb{E}_{q} 
    \Big{[}\log p(\vx_0 | \vx_1) - \mathrm{KL} \big{(}  q(\vx_T | \vx_0) | p(\vx_T) \big{)} - \sum_{t=2}^T \mathrm{KL} \big{(}q(\vx_{t-1} | \vx_t, \vx_0) | p(\vx_{t-1} | \vx_t)\big{)} \Big{]}
\end{split}
\label{eq:diffusion_final_objective}
\end{align}\normalsize

\section{Argmax Flows}
Argmax flows define discrete distributions using 1) a density model $p(\vv)$, such as a normalizing flow, and 2) an argmax layer that maps the continuous $\vv \in \bR^{D \times K}$ to a discrete $\vx \in \{1,2,...,K\}^D$ using 
\begin{equation}
\small
    \vx = \argmax \vv \quad \text{ where } \quad x_d = \argmax_k v_{dk}.
\end{equation}
This is a natural choice to model categorical variables, because it divides the entire continuous space of $\vv$ into symmetric partitions corresponding to categories in $\vx$. To sample from an argmax flow sample $\vv \sim p(\vv)$ and compute $\vx = \argmax \vv$ (Algorithm~\ref{alg:sample_argmax}). To generate reasonable samples, it is up to the density model $p(\vv)$ to capture any complicated dependencies between the different dimensions. While sampling from an argmax flow is straightforward, the main difficulty lies in \textit{optimizing} this generative model. To compute the likelihood of a datapoint $\vx$, we have to compute
\begin{equation}
\small
    P(\vx) =\! \int P(\vx | \vv) p(\vv)d\vv, ~~P(\vx|\vv) \!=\! \delta\big(\vx \!=\! \argmax(\vv)\big),
\end{equation}
which is intractable. Consequently, we resort to variational inference and specify a variational distribution $q(\vv|\vx)$. We note that na\"{i}vely choosing any variational distribution may lead to samples $\vv \sim q(\vv | \vx)$ where $\delta(\vx = \argmax \vv) = 0$, which yields an ELBO of negative infinity. To avoid this, we need a variational distribution $q(\vv|\vx)$ that satisfies what we term the \emph{argmax constraint}:
\begin{equation*}
\small
    \vx = \argmax \vv \quad \text{for all} \quad \vv \sim q(\vv|\vx).
\end{equation*}
That is, the variational distribution $q(\vv|\vx)$ should have support limited to $\mathcal{S}(\vx) = \{\vv \in \mathbb{R}^{D\times K} : \vx = \argmax \vv\}$. Recall that under this condition, the ELBO simplifies to $\Eb_{\vv \sim q(\vv|\vx)} \left[ \log p(\vv) - \log q(\vv|\vx) \right]$, as shown in Algorithm~\ref{alg:optimize_argmax}. For an illustration of the method see Figure \ref{fig:overview_argmax}.

\begin{table}
% \begin{minipage}[t]{.99\textwidth}
\begin{minipage}[t]{.47\textwidth}
\begin{algorithm}[H]
   \caption{Sampling from Argmax Flows}
   \label{alg:sample_argmax}
\begin{algorithmic}
\STATE {\bfseries Input:} $p(\vv)$
   \STATE {\bfseries Output:} Sample $\vx$
\STATE Sample $\vv \sim p(\vv)$
\STATE Compute $\vx = \argmax \vv$
\end{algorithmic}
\end{algorithm}
\end{minipage}
\hfill
\begin{minipage}[t]{.47\textwidth}
\begin{algorithm}[H]
   \caption{Optimizing Argmax Flows}
   \label{alg:optimize_argmax}
\begin{algorithmic}
   \STATE {\bfseries Input:} $\vx$, $p(\vv)$, $q(\vv|\vx)$
   \STATE {\bfseries Output:} ELBO $\mathcal{L}$
\STATE Sample $\vv \sim q(\vv|\vx)$
\STATE Compute $\mathcal{L} = \log p(\vv) - \log q(\vv|\vx)$
\end{algorithmic}
\end{algorithm}
\end{minipage}
% \end{minipage}
\end{table}

\subsection{Probabilistic Inverse}

The argmax layer may be viewed as a surjective flow layer \citep{nielsen2007survae}. With this view, the variational distribution $q(\vv|\vx)$ specifies a distribution over the possible right-inverses of the argmax function, also known as a \emph{stochastic inverse} or \emph{probabilistic inverse}. 
Recall that the commonly-used dequantization layer for ordinal data corresponds to the probabilistic inverse of a rounding operation. As summarized in Table \ref{tab:surjective}, this layer may thus be viewed as analogous to the argmax layer, where the round is for ordinal data while the argmax is for categorical data.

We are free to specify any variational distribution $q(\vv|\vx)$ that satisfies the argmax constraint. In the next paragraphs we outline three possible approaches. Since operations are performed independently across dimensions, we omit the dimension axis and let $\vv \in \mathbb{R}^K$ and $x \in \{1, \ldots, K\}$.

\paragraph{Thresholding (Alg.~\ref{alg:thresholding}).} 
A straightforward method to construct a distribution $q(\vv|x)$ satisfying the argmax constraint is to use thresholding. That is, we first sample an unbounded variable $\vu \in \mathbb{R}^{K}$ from $q(\vu | x)$, which can be for example a conditional Gaussian or normalizing flow. Next, we map $\vu$ to $\vv$ such that element $x$ is the largest:
\begin{equation}
    v_x = u_x \quad \text{and} \quad \vv_{-x} = \mathrm{threshold}_T(\vu_{-x})
\end{equation}
where the thresholding is applied elementwise with threshold value $T=v_x$. This ensures that element $v_x$ is the largest, and consequently that $q(\vv|x)$ satisfies the argmax constraint. 
Note that we require the threshold function to be bijective, $\mathrm{threshold}_T: \bR \rightarrow (-\infty,T)$, so that we can use the change-of-variables formula to compute $\log q(\vv|x)$.
In our implementation, thresholding is implemented using a softplus such that all values are mapped below a limit $T$:
\begin{equation}
\small
   v = \mathrm{threshold}_T(u) = T-\mathrm{softplus}(T-u), 
\end{equation}
where $\mathrm{softplus}(z) = \log(1 + e^z)$ and for which it is guaranteed that $v \in (-\infty, T)$.

\begin{table}[]
\begin{minipage}[t]{.485\textwidth}
\begin{algorithm}[H]
   \caption{Thresholding-based $q(\vv|\vx)$}
   \label{alg:thresholding}
\begin{algorithmic}
   \STATE {\bfseries Input:} $\vx$, $q(\vu|\vx)$
   \STATE {\bfseries Output:} $\vv$, $\log q(\vv|\vx)$
\STATE $\vu \sim q(\vu|\vx)$
\STATE $\vv_{\vx} = \vu_{\vx}$
\STATE $\vv_{-\vx} = \mathrm{threshold}(\vu_{-\vx}, \vx)$
\STATE $\log q(\vv|\vx) = \log q(\vu|\vx) - \log |\det \rmd\vv / \rmd\vu|$
\end{algorithmic}
\end{algorithm}
\end{minipage}
\hfill
\begin{minipage}[t]{.485\textwidth}
\begin{algorithm}[H]
   \caption{Gumbel-based $q(\vv|\vx)$}
   \label{alg:gumbel}
\begin{algorithmic}
   \STATE {\bfseries Input:} $\vx$, $\vphi$
   \STATE {\bfseries Output:} $\vv$, $\log q(\vv|\vx)$
% \STATE $\vphi \leftarrow \mathrm{NN}(\vx)$
\STATE $\phi_{\max} = \log \sum_i \exp \phi_i$
\STATE $\vv_{\vx} \sim \mathrm{Gumbel}(\phi_{\max})$
\STATE $\vv_{-\vx} \sim \mathrm{TruncGumbel}(\vphi_{-\vx}, \vv_{\vx})$
\STATE $\log q(\vv|\vx) = \log \mathrm{Gumbel}(\vv_\vx | \phi_{\max})$
\STATE \hspace{1.0cm} $+ \log \mathrm{TruncGumbel}(\vv_{-\vx} | \vphi_{-\vx}, \vv_\vx)$
\end{algorithmic}
\end{algorithm}
\end{minipage}
\end{table}

\paragraph{Gumbel (Alg.~\ref{alg:gumbel}).} 
An alternative approach is to let $q(\vv|x) = \mathrm{Gumbel}(\vv|\vphi)$ \textit{restricted} to $\argmax \vv = x$, where the location parameters $\vphi \leftarrow \mathrm{NN}(x)$ are predicted using a neural network $\mathrm{NN}$. 
The Gumbel distribution has favourable properties: The $\argmax$ and $\max$ are independent and the $\max$ is also distributed as a Gumbel:
\begin{equation}
\small
\label{eq:gumbel_max}
    \max_i v_i \sim \mathrm{Gumbel}(\phi_{\max}), 
\end{equation}
where $\phi_{\max} = \log \sum_i \exp \phi_i$.
For a more extensive introduction see \citep{Maddison2014Astarsampling,kool2019}. To sample $\vv \sim q(\vv|x)$, we thus first sample the maximum $v_x$ according to Eq.~\ref{eq:gumbel_max}. Next, given the sample $v_x$, the remaining values can be sampled using \textit{truncated} Gumbel distributions:
\begin{equation}
\small
    v_{i} \sim \mathrm{TruncGumbel}(\phi_i; T) \text{ where } i \not= x
    \label{eq:gumbel_remaining}
\end{equation}
where the truncation value $T$ is given by $v_x$ which ensures that the argmax constraint $v_x > v_i$ for $i\not=x$ is satisfied. Recall that to optimize Eq.~\ref{eq:elbo_objective}, $\log q(\vv | \vx)$ is also required, which can be computed using the closed-form expressions for the log density functions (see Table \ref{tab:gumbel}). Another property of Gumbel distributions is that 
\begin{equation}
\small
    P(\argmax \vv = i) = \exp \phi_i / \sum_i \exp \phi_i,
\end{equation}
which we use to initialize the location parameters $\boldsymbol{\phi}$ to match the empirical distribution of the first minibatch of the data. 

\paragraph{Gumbel Thresholding.}  This method unifies the methods from the previous two sections: Gumbel distributions and thresholding. The key insight is that the Gumbel sampling procedures as defined above can be seen as a reparametrization of a uniform noise distribution $\mathcal{U}(0, 1)^K$ which is put through the inverse CDF of the Gumbel distributions (see Table \ref{tab:gumbel}). From the perspective of change-of-variables, the log likelihood denotes the log volume change of this transformation. To increase expressitivity the uniform distribution can be replaced by a normalizing flow $q(\vu | x)$ that has support on the interval $(0, 1)^K$, which can be enforced using a sigmoid transformation. This section shows that a large collection of thresholding functions can be found by studying (truncated) inverse CDFs. In practice we find that performance is reasonably similar as long as the underlying noise $\vu$ is learned.

\paragraph{Behavior of the Variational Posterior}
Although several methods to learn $q$ have been proposed, it is unclear what expressitivity is required. In the following, the interactions between $q(\vv | \vx)$ and the density model $p(\vv)$ are discussed. Recall that the variational bound that is optimized under expectation of a data distribution $\mathcal{D}$ can be seen as minimizing the $\operatorname{KL}$ distance between the aggregated posterior $q(\vv) = \mathbb{E}_{\vx \sim \mathcal{D}}q(\vv | \vx)$ and the density model $p(\vv)$, so $\operatorname{KL}(q(\vv)|p(\vv))$. There are two distinct reasons which can cause this distance to be large: Firstly, the density model $p(\vv)$ may not have the right probability mass in each argmax region. These desired probabilities solely depend on the data distribution $\mathcal{D}$. Secondly, the variational posterior $q(\vv | \vx)$ may not have the correct shape compared to $p(\vv)$, \textit{within} an argmax region. At initialization, the thresholding within $q$ can create low density regions at argmax boundaries.

In theory, if $p(\vv)$ is a universal density approximator, then the model can be fitted for any well-behaved $q(\vv|\vx)$. Then $p(\vv)$ can even fit the low density regions in the boundaries. This argument is trivial, as one can simply set $p(\vv)$ to $q(\vv) = \mathbb{E}_{\vx \sim \mathcal{D}}q(\vv | \vx)$. In practice, over training steps we find that $q$ does smooth out these boundary artifacts, and counteracts the thresholding so that the aggregated posterior becomes smoother.

\subsection{Cartesian Products of Argmax Flows}
In the current description, Argmax Flows require the same number of dimensions in $\vv$ as there are classes in $\vx$. To alleviate this constraint we introduce Cartesian products of Argmax Flows. To illustrate our method, consider a 256 class problem. One class can be represented using a single number in $\{1, \ldots, 256\}$, but also using two hexadecimal numbers $\{1, \ldots, 16\}^2$ or alternatively using eight binary numbers. Specifically, any base $K$ variable $\vx^{(K)} \in \{1, \ldots, K\}^D$ can be converted to a base $M$ variable $\vx^{(M)} \in \{1, \ldots, M\}^{d_m \times D}$ where $d_m = \lceil \log_M K \rceil$. 
Then the variable $\vx^{(M)}$ with dimensionality $M \cdot d_m \cdot D$ represents the variable $\vx^{(K)}$ with dimensionality $K \cdot D$, trading off symmetry for dimensionality. Even though this may lead to some unused additional classes, the ELBO objective in Equation \ref{eq:elbo_objective} can still be optimized using an $M$-categorical Argmax Flow. Finally, note that Cartesian products of binary spaces are a special case where the variable can be encoded symmetrically into a single dimension to the positive and negative part using binary dequantization \citep{winkler2019learning}. In this case, by trading-off symmetry the dimensionality increases only proportional to $\log_2 K$ .

\section{Multinomial Diffusion}
In this section we introduce an alternative likelihood-based model for categorical data: Multinomial Diffusion. In contrast with previous sections, $\vx_t$ will be represented in one-hot encoded format $\vx_t \in \{0, 1\}^K$. Specifically, for category $k$, $x_{k} = 1$ and $x_{j} = 0$ for $j \not= k$. Note that again the dimension axis is omitted for clarity as all distributions are independent over the dimension axis. We define the multinomial diffusion process using a categorical distribution that has a $\beta_t$ chance of resampling a category uniformly:
\begin{equation}
    q(\vx_t | \vx_{t-1}) =  \mathcal{C}(\vx_t | (1 - \beta_t) \vx_{t-1} + \beta_t / K ),
    \label{eq:diff_categorical_forward}
\end{equation}
where $\mathcal{C}$ denotes a categorical distribution with probability parameters after $|$. Further addition (and subtraction) between scalars and vectors is done elementwise. This convention kept throughout this section.
Since these distributions form a Markov chain, we can express the probability of any $\vx_t$ given $\vx_0$ as:
\begin{equation}
    q(\vx_t | \vx_{0}) =  \mathcal{C}(\vx_t | \bar{\alpha}_t \vx_{0} + (1 - \bar{\alpha}_t) / K )
    \label{eq:diff_categorical_forward_x0}
\end{equation}
where $\alpha_t = 1 - \beta_t$ and $\bar{\alpha}_t = \prod_{\tau=1}^t \alpha_\tau$. Intu\"{i}tively, for each next timestep, a little amount of uniform noise $\beta_t$ over the $K$ classes is introduced, and with a large probability $(1 - \beta_t)$ the previous value $\vx_{t-1}$ is sampled. Using Equation \ref{eq:diff_categorical_forward} and \ref{eq:diff_categorical_forward_x0} the categorical posterior $q(\vx_{t-1} | \vx_{t}, \vx_0)$ can be computed in closed-form:
\begin{align}
\begin{split}
    q(\vx_{t-1} | \vx_{t}, \vx_0) &= \mathcal{C}(\vx_{t-1} | \vtheta_{\mathrm{post}}(\vx_t, \vx_0)), \,\, \text{ where } \,\, \vtheta_{\mathrm{post}}(\vx_t, \vx_0) = \tilde{\vtheta} / \sum_{k=1}^K \tilde{\theta}_k
     \\
    \text{and } \,\tilde{\vtheta} &= [\alpha_t \vx_t + (1 - \alpha_t) / K] \odot [\bar{\alpha}_{t-1} \boldsymbol{x}_0 + (1 - \bar{\alpha}_{t-1}) / K ].
    \label{eq:q_posterior}
\end{split}
\end{align}

\begin{figure}
\centering
% {\includegraphics[width=0.5\linewidth]{images/nodes}}
    \includegraphics[width=.99\linewidth]{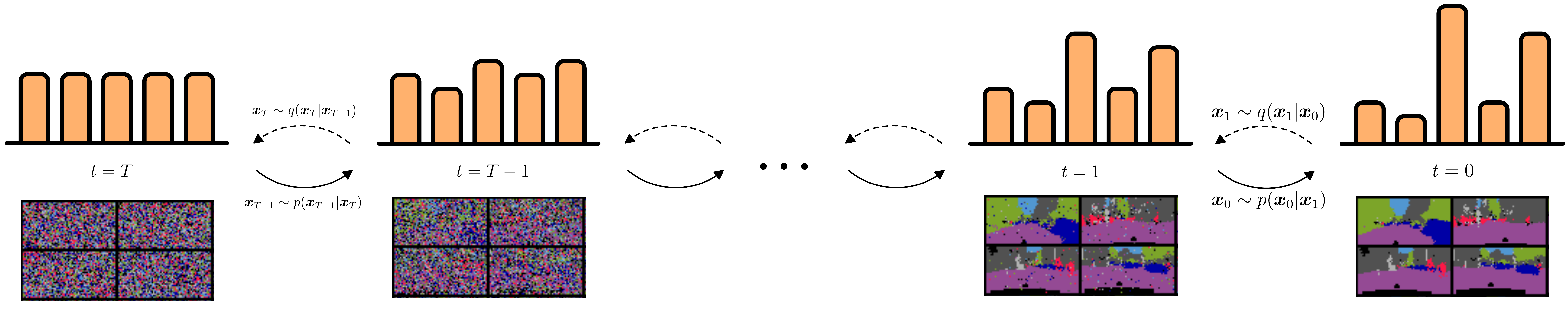}
    \caption{Overview of multinomial diffusion. A generative model $p(\vx_{t-1} | \vx_{t})$ learns to gradually denoise a signal from left to right. An inference diffusion process $q(\vx_{t} | \vx_{t-1})$ gradually adds noise form right to left.}
    \label{fig:overview_mult_diffusion}
\end{figure}
One of the innovations in \cite{ho2020denoising} was the insight to not predict the parameters for the generative trajectory directly, but rather to predict the noise using the posterior equation for $q$. Although predicting the noise is difficult for discrete data, we predict a probability vector for $\hat{\vx}_0$ from $\vx_t$ and subsequently parametrize $p(\vx_{t-1} | \vx_{t})$ using the probability vector from $q(\vx_{t-1} | \vx_t, \hat{\vx}_0)$, where $\vx_0$ is approximated using a neural network $\hat{\vx}_0 = \mu(\vx_t, t)$. Equation~\ref{eq:q_posterior} will produce valid probability vectors that are non-negative and sums to one under the condition that the prediction $\hat{\vx}_0$ is non-negative and sums to one, which is ensured with a softmax function in $\mu$. To summarize:
\begin{align}
    \begin{split}
    p(\vx_{0} | \vx_{1}) = \mathcal{C}(\vx_{0} | \hat{\vx}_0) \, \text{ and } \,
    p(\vx_{t-1} | \vx_{t}) = \mathcal{C}(\vx_{t-1} | \boldsymbol{\theta}_{\mathrm{post}}(\vx_t, 
    \hat{\vx}_0)) \, 
    \text{ where }\, \hat{\vx}_0 = \mu(\vx_t, t)
    \end{split}
    \label{eq:p_parametrization}
\end{align}
The KL terms in Equation \ref{eq:diffusion_final_objective} can be simply computed by enumerating the probabilities in Equation \ref{eq:q_posterior} and \ref{eq:p_parametrization} and computing the KL divergence for discrete distributions in $L_{t-1}$ with $t \geq 2$:
\small\begin{align}
\begin{split}
    \mathrm{KL} \big{(}q(\vx_{t-1} | \vx_t, \vx_0) | p(\vx_{t-1} | \vx_t)\big{)}  &= \mathrm{KL} \big{(} \mathcal{C}(\boldsymbol{\theta}_{\mathrm{post}}(\vx_t, \vx_0)) | \mathcal{C}(\boldsymbol{\theta}_{\mathrm{post}}(\vx_t, 
    \hat{\vx}_0))\big{)},
\end{split}
\end{align}\normalsize
which can be computed using $\sum_k \boldsymbol{\theta}_{\mathrm{post}}(\vx_t, \vx_0))_k \cdot \log \frac{\boldsymbol{\theta}_{\mathrm{post}}(\vx_t, \vx_0))_k}{\boldsymbol{\theta}_{\mathrm{post}}(\vx_t, 
\hat{\vx}_0))_k}$. Furtermore, to compute $\log p(\vx_0 | \vx_1)$ use that $\vx_0$ is onehot:
\small\begin{equation}
    \log p(\vx_0 | \vx_1) = \sum_k \vx_{0,k} \log  \hat{\vx}_{0,k}
\end{equation}\normalsize

\section{Related Work}

Deep generative models broadly fall into the categories autoregressive models ARMs \citep{germain2015made}, Variational Autoencoders (VAEs) \citep{kingma2014auto,rezende2014stochasticvariationalinference}, Adversarial Network (GANs) \citep{goodfellow2014generative}, Normalizing Flows \citep{rezende2015norm}, Energy-Based Models (EBMs) and Diffusion Models \citep{sohl2015diffusion}.

Normalizing Flows typically learn a continuous distribution and dequantization is required to train these methods on ordinal data such as images. A large body of work is dedicated to building more expressive continuous normalizing flows \citep{dinh2016density,germain2015made,kingma2016improved,papamakarios2017masked,chen2018neural,song2019mintnet,perugachi2020idensenets}. 
To learn ordinal discrete distributions with normalizing flows, adding uniform noise in-between ordinal classes was proposed in \citep{uria2013rnade} and later theoretically justified in \citep{theis2016note}. An extension for more powerful dequantization based on variational inference was proposed in \citep{ho2019flow++}, and connected to autoregressive models in \citep{nielsen2020closingdequantizationgap}. Dequantization for binary variables was proposed in \citep{winkler2019learning}. \citet{discrete2019tran} propose invertible transformations for categorical variables directly. However, these methods can be difficult to train because of gradient bias and results on images have thus far not been demonstrated. In addition flows for ordinal discrete data (integers) have been explored in \citep{hoogeboom2019integer,berg2020idfpp}. In other works, VAEs have been adapted to learn a normalizing flow for the latent space \citep{ziegler2019latent, lippe2020categorical}. However, these approaches typically still utilize an argmax heuristic to sample, even though this is not the distribution specified during training. 

Diffusion models were first introduced in \cite{sohl2015diffusion}, who developed diffusion for Gaussian and Bernoulli distributions. Recently, Denoising Diffusion models \cite{ho2020denoising} have been shown capable of generating high-dimensional images by architectural improvements and reparametrization of the predictions. Diffusion models are relatively fast to train, but slow to sample from as they require iterations over the many timesteps in the chain. \citet{song2020denoisingimplicit,nichol2021improved} showed that in practice samples can be generated using significantly fewer steps. \citet{nichol2021improved} demonstrated that importance-weighting the objective components greatly improves log-likelihood performance. In \cite{song2020scorebasedSDEs} a continuous-time extension of denoising diffusion models was proposed. After initial release of this paper we discovered that \citet{song2020denoisingimplicit} concurrently also describe a framework for discrete diffusion, but without empirical evaluation.

\section{Experiments}
\label{sec:experiments}
In our experiments we compare the performance of our methods on language modelling tasks and learning image segmentation maps unconditionally.

\begin{table}
    \centering
    \caption{Comparison of a coupling and autoregressive generative flows with uniform \citep{uria2013rnade} and variational \citep{ho2019flow++} dequantization and our proposed Argmax flows.}
    \label{tab:performance_generative_flow_text}
    \scalebox{.9}{
    \begin{tabular}{l l c c c c}
    \toprule
    Dequantization & Flow type & text8 (bpc) & enwik8 (bits per raw byte) \\ \midrule
    Uniform dequantization  & \multirow{3}{*}{Autoregressive} & 1.90 & 2.14 \\
    Variational dequantization  &  & 1.43 & 1.44 \\
    Argmax Flow (ours) & & \textbf{1.38} & \textbf{1.42} \\ \midrule
    Uniform dequantization &  \multirow{3}{*}{Coupling} & 2.01 & 2.33 \\
    Variational dequantization & & 2.08 & 2.28 \\
    Argmax Flow (ours) & & \textbf{1.82} & \textbf{1.93} \\ \midrule
    \end{tabular}}
\vspace{-.3cm}
\end{table}

\begin{table}[b]
    \centering
    \caption{Comparison of different methods on \texttt{text8} and \texttt{enwik8}. Results are reported in negative log-likelihood with units bits per character (bpc) for \texttt{text8} and bits per raw byte (bpb) for \texttt{enwik8}.}
    \label{tab:performance_text}
    \scalebox{.9}{
    \begin{tabular}{l l l c c c}
    \toprule
    Model type & & Model & text8 (bpc) & enwik8 (bpb) \\ \midrule
    \multirow{2}{*}{ARM} 
    % & &  LSTM {\footnotesize \citep{Cooijmans2017recurrentbatch}} & 1.42 & - & 19.8$^\dagger$ \\
    % & &  Large mLSTM {\footnotesize \citep{krause2017multiplicativelstm}} & 1.27 & 1.24 & - \\
    & & 64 Layer Transformer {\footnotesize \citep{alrfou2019transformer}} & 1.13 & 1.06 \\
    & & TransformerXL {\footnotesize \citep{dai2019transformerxl}} & 1.08 & 0.99 \\ \midrule \midrule
    % & & Sparse Transformer {\footnotesize \citep{child2019sparse}} & - & 0.99 & - \\ \cline{3-6}
   \multirow{3}{*}{\small VAE} & & AF/AF$^\star$ (AR) {\footnotesize \citep{ziegler2019latent}} & 1.62 & 1.72\\
   & & IAF / SCF$^\star$ {\footnotesize \citep{ziegler2019latent}} & 1.88 & 2.03 \\
    & & CategoricalNF (AR) {\footnotesize \citep{lippe2020categorical}} & 1.45 & - \\ \midrule
    \multirow{2}{*}{Generative Flow}
    & & Argmax Flow, AR (ours) & 1.39 & 1.42 \\
    & & Argmax Coupling Flow (ours) & 1.82 & 1.93 \\
    \midrule
    Diffusion & & Multinomial Text Diffusion (ours) & 1.72 & 1.75 \\ \bottomrule
    \end{tabular}}
\begin{flushleft}
 \scriptsize{$\star$ Results obtained by running code from the official repository for the \texttt{text8} and \texttt{enwik8} datasets.} \\
 \vspace{-.5cm}
\end{flushleft}
\end{table}

\subsection{Language data}
In this section we compare our methods on two language datasets,  \texttt{text8} and \texttt{enwik8}. \texttt{text8} contains 27 categories (`a' through `z' and ` ') and for \texttt{enwik8} the bytes are directly modelled which results in 256 categories.

\paragraph{Model description}
Two versions of generative argmax flows are tested: using an autoregressive (AR) flow and a coupling-based flow for $p(\vv)$. In these experiments the probabilistic inverse is based on the thresholding approach. Specifically, a conditional diagonal Gaussian $q(\vu | \vx)$ is trained and thresholded which gives the distribution $q(\vv | \vx)$.
The argmax flow is defined on binary Cartesian products. This means that for $K=27$, a $5$-dimensional binary space is used and for $K=256$ an $8$-dimensional binary space. The argmax flow is compared to the current standard of training generative flows directly on discrete data: dequantization. We compare to both uniform and variational dequantization, where noise on a (0, 1) interval is added to the onehot representation of the categorical data. The autoregressive density model is based on the model proposed in \citep{lippe2020categorical}. The coupling density model consists of 8 flow layers where each layer consists of a $1$ $\times$ $1$ convolution and mixture of logistics transformations \cite{ho2019flow++}. In the multinomial text diffusion model, the $\mu$ network is modeled by a 12-layer Transformer. For more extensive details about the experiment setup see Appendix \ref{app:experimental_details}.

\paragraph{Comparison with Generative Flows} 
% Models are categorized as autoregressive or non-autoregressive. 
Firstly we compare the performance of generative flows directly trained on language data (Table~\ref{tab:performance_generative_flow_text}). These experiments are using the same underlying normalizing flow: either a coupling-based flow or an autoregressive flow. Note that Argmax Flows consistently outperform both uniform and variational dequantization. This indicates that it is easier for a generative flow to learn the lifted continuous distribution using an argmax flow. An advantage of Argmax flows that may explain this difference is that they lift the variables into the entire Euclidean space, whereas traditional dequantization only introduce probability density on $(0, 1)$ intervals, leaving gaps with no probability density. The performance improvements of Argmax flows are even more pronounced when comparing coupling-based approaches. Also note that coupling flows have worse performance than autoregressive flows, with a difference that is generally smaller for images. This indicates that designing more expressive coupling layers for text is an interesting future research direction.

\paragraph{Comparison with other generative models} The performance compared to models in literature is presented in Table~\ref{tab:performance_text} alongside the performance of our Argmax Flows and Multinomial Diffusion. The latent variable approaches containing autoregressive components are marked using (AR). Although autoregressive flows still have the same disadvantages as ARMs, they provide perspective on where performance deficiencies are coming from. We find that our autoregressive Argmax Flows achieve better performance than the VAE approaches, they outperform AF/AF \citep{ziegler2019latent} and CategoricalNF \citep{lippe2020categorical}.

\begin{figure}
\begin{minipage}[t]{.48\textwidth}
% \vspace{.3cm}
\begin{figure}[H]
\centering
\begin{subfigure}{.99\textwidth}
\includegraphics[width=\textwidth]{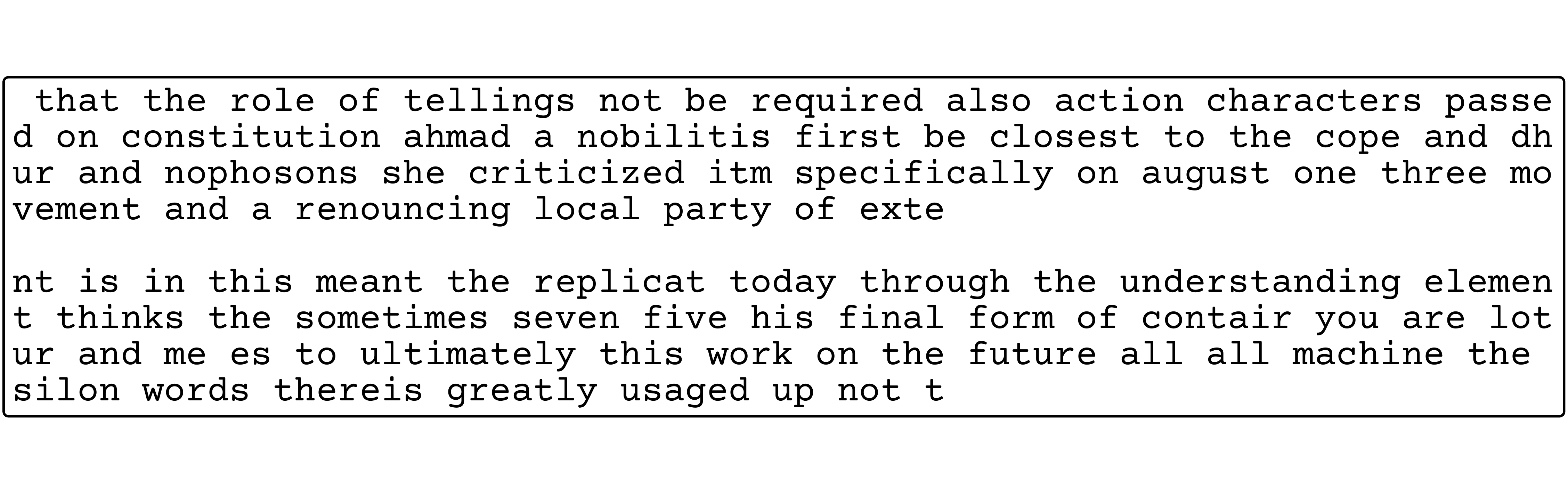}
\caption{Samples from Multinomial Text Diffusion.}
% \vspace{.65cm}
\end{subfigure}
\begin{subfigure}{.99\textwidth}
\includegraphics[width=\textwidth]{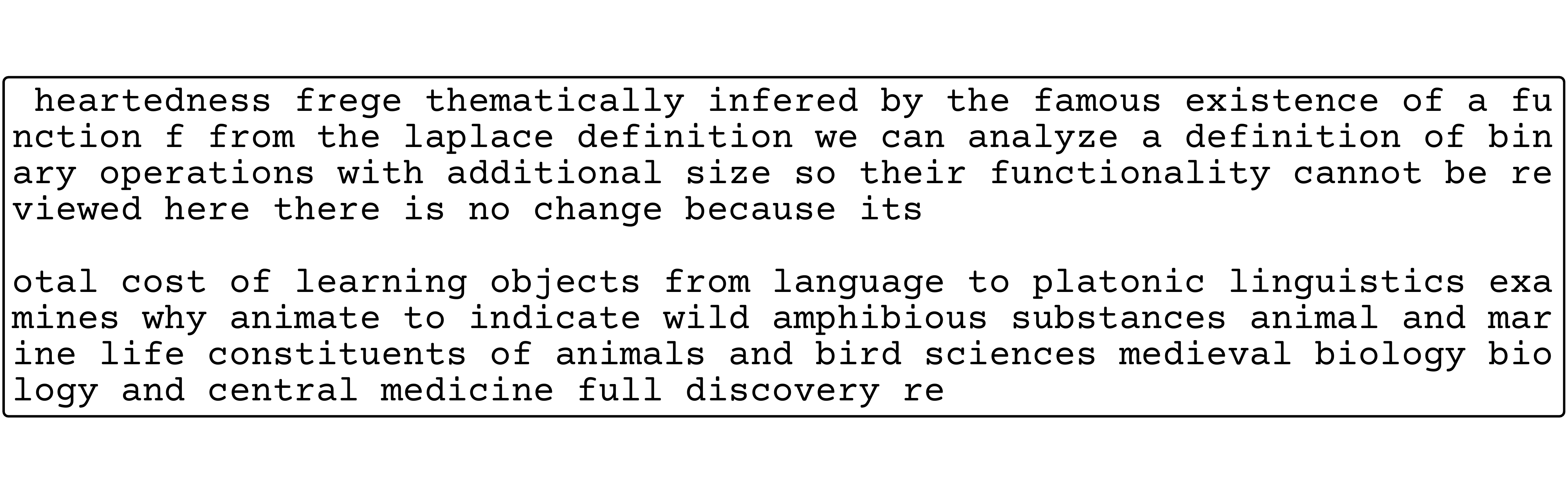}
\caption{Samples from Argmax AR Flow.}
% \vspace{.65cm}
\end{subfigure}
\begin{subfigure}{.99\textwidth}
\includegraphics[width=\textwidth]{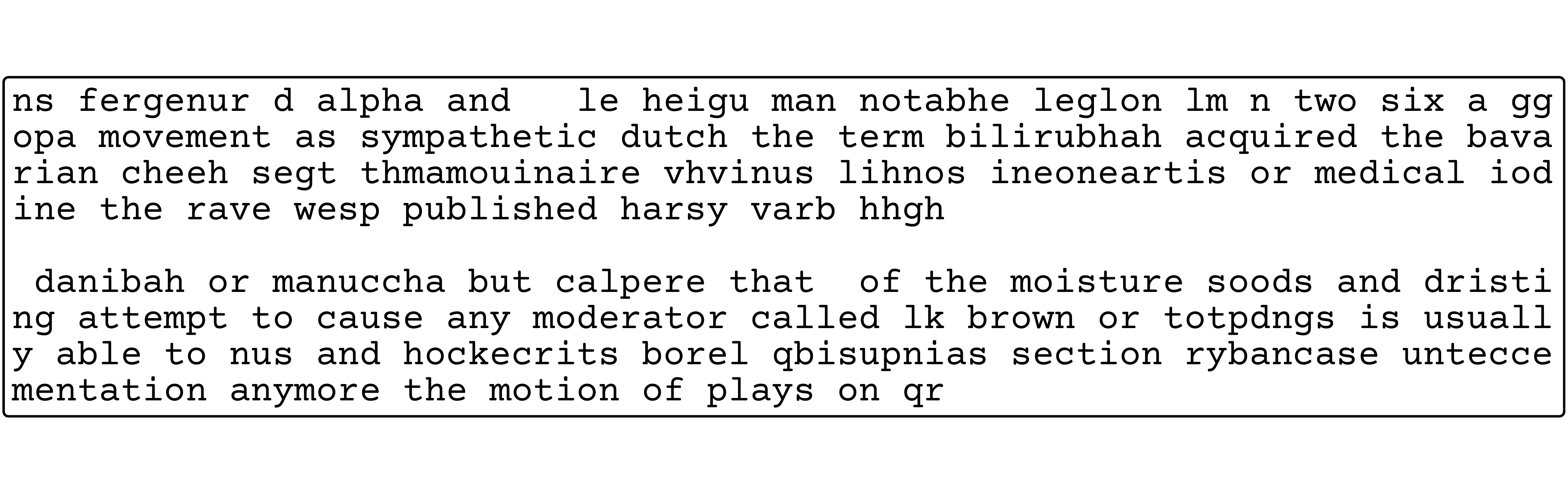}
\caption{Samples from Argmax Coupling Flow.}
\end{subfigure}
\caption{Samples from models, \texttt{text8}.}
\label{fig:samples_text}
\end{figure}
\end{minipage}
\hfill
\begin{minipage}[t]{.48\textwidth}
\begin{figure}[H]
\centering
\begin{subfigure}{.99\textwidth}
\includegraphics[width=\textwidth]{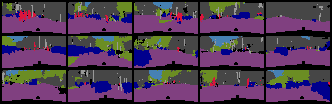}
\caption{Samples from the Argmax Flow.}
\end{subfigure}
\begin{subfigure}{.99\textwidth}
\includegraphics[width=\textwidth]{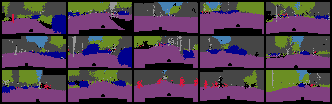}
\caption{Samples from the Multinomial Diffusion model.}
\end{subfigure}
\begin{subfigure}{.99\textwidth}
\includegraphics[width=\textwidth]{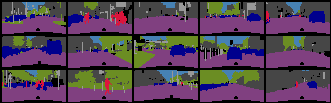}
\caption{Cityscapes data.}
\end{subfigure}
\caption{Samples from models, cityscapes.}
\label{fig:experiments_cityscapes}
\vspace{-.5cm}
\end{figure}
\end{minipage}
\end{figure}

When comparing non-autoregressive models, Argmax Flows also outperforms the method that lifts the categorical space to a continuous space: IAF / SCF \citep{ziegler2019latent}. Interestingly, the multinomial text diffusion is a non-autoregressive model that performs even better than the argmax coupling flow, but performs worse than the autoregressive version. For this model it is possible that different diffusion trajectories for $q$ would result in even better performance, because in the current form the denoising model has to be very robust to input noise. These experiments also highlight that there is still a distinct performance gap between standard ARMs and (autoregressive) continuous density model on text, possibly related to the dequantization gap \citep{nielsen2020closingdequantizationgap}. Samples from different models trained on \texttt{text8} are depicted in Figure~\ref{fig:samples_text}. Because of difficulties in reproducing results from Discrete Flows, a comparison and analysis of discrete flows are left out of this section. Instead they are extensively discussed in Appendix~\ref{sec:reproducing_discrete_flows}. For additional experiments regarding Cartesian products and sampling time see Appendix~\ref{sec:additional_experiments}.

\begin{wrapfigure}{r}{.5\textwidth}
\begin{minipage}[t]{.485\textwidth}
\vspace{-.2cm}
\begin{figure}[H]
\centering
\begin{subfigure}{.99\textwidth}
\includegraphics[width=\textwidth]{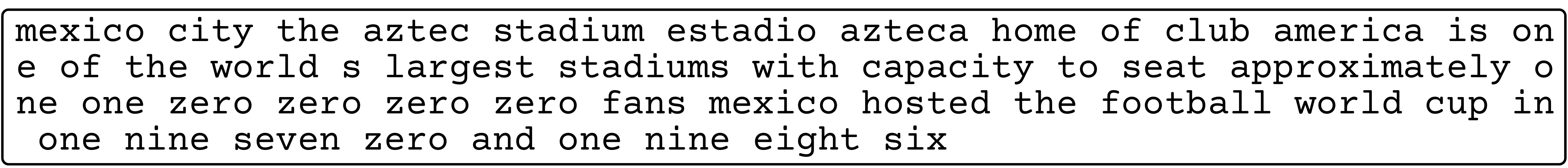}
\caption{Ground truth sequence from \texttt{text8}.}
\end{subfigure}
\begin{subfigure}{.99\textwidth}
\includegraphics[width=\textwidth]{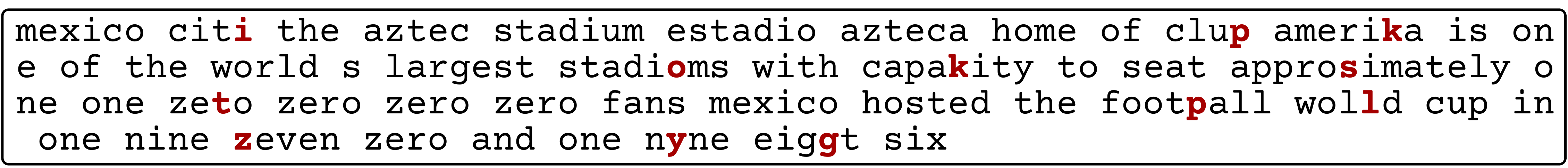}
\caption{Corrupted sentence.}
\end{subfigure}
\begin{subfigure}{.99\textwidth}
\includegraphics[width=\textwidth]{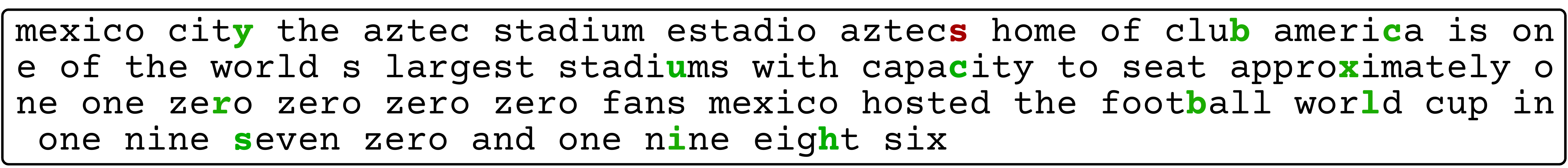}
\caption{Suggested, prediction by the model.}
\end{subfigure}
\caption{Spell checking with Multinomial Text Diffusion.}
\label{fig:spellcheck}
\end{figure}
\end{minipage}
\vspace{-.3cm}
\end{wrapfigure}

\paragraph{Unsupervised spell-checking}
An interesting by-product of the text diffusion model is that it can be used to spell-check text using a single forward pass. To demonstrate this, a sentence taken from the test data is corrupted by changing a few characters. This corrupted sequence is given as $\vx_1$ to the generative denoising model, which is close to the data at step $0$. Then the denoising model predicts $p(\vx_0 | \vx_1)$ and the most-likely $\vx_0$ can be suggested. Note that this model only works for character-level corruption, not insertions. An example is depicted in Figure~\ref{fig:spellcheck}. Since the model chooses the most-likely matching word, larger corruptions will at some point lead to word changes.

\subsection{Segmentation maps}
For image-type data, we introduce a categorical image dataset: the cityscapes dataset is repurposed for \textit{unconditional} image segmentation learning. In contrast with the standard setting, the distribution over the segmentation targets needs to be learned \textit{without} conditioning on the photograph. To reduce computational cost, we rescale the segmentation maps from cityscapes to $32 \times 64$ images using nearest neighbour interpolation. We utilize the global categories as prediction targets which results in an 8-class problem.

\begin{wrapfigure}{r}{.5\textwidth}
\begin{minipage}[t]{.49\textwidth}
\vspace{-1.1cm}
 \begin{table}[H]
    \centering
    \caption{Performance of different dequantization methods on squares and cityscapes dataset, in bits per pixel, lower is better.}
    \label{tab:results_cityscapes}
    % \label{tab:my_label}
    \scalebox{.89}{
    \begin{tabular}{l l r r r r}
    \toprule
    Cityscapes & ELBO & IWBO  \\ \midrule
    Round / Unif. \citep{uria2013rnade} & 1.010 & 0.930 \\
    Round / Var. \citep{ho2019flow++} &  0.334 & 0.315 \\ \midrule
    % Discrete Flow &  \\
    % VAE (Flow prior, Flow var. posterior) & 0.306 (0.293) \\ \midrule
    Argmax / Softplus thres. (ours) & \textbf{0.303} & \textbf{0.290} \\
    Argmax / Gumbel dist. (ours) & 0.365 & 0.341 \\
    Argmax / Gumbel thres. (ours) & \textbf{0.307} & \textbf{0.287} \\ \midrule
    % 0.3074 (0.2857) 0.3060 0.2873
    Multinomial Diffusion (ours) & \multicolumn{2}{c}{0.305} \\ \bottomrule
   \end{tabular}}
    \vspace{-.1cm}
\end{table}
\end{minipage}
\vspace{-.3cm}
\end{wrapfigure}

\vspace{-.15cm}
\paragraph{Model description}
The Argmax Flows are defined directly on the $K=8$ categorical space. The density model $p(\vv)$ is defined using affine coupling layers parametrized by DenseNets \citep{huang2017densely}. For the probabilistic inverse we learn a conditional flow $q(\vu | \vx)$ which is also based on the affine coupling structure. Depending on the method, either softplus or Gumbel thresholding is applied to obtain $\vv$. Recall that for our first Gumbel approach it is equivalent to set $q(\vu | \vx)$ to the unit uniform distribution, whereas $q(\vu | \vx)$ is learned for Gumbel thresholding. We compare to existing dequantization strategies in literature: uniform \citep{uria2013rnade} and variational dequantization \citep{ho2019flow++} which are applied on the onehot representation. All models utilize the same underlying flow architectures and thus the number of parameters is roughly the same. The exception are uniform dequantization and the Gumbel distribution, since no additional variational flow distribution is needed. For more extensive details see Appendix \ref{app:experimental_details}.

\vspace{-.15cm}
\paragraph{Comparison}
The results of this experiment are shown in Table~\ref{tab:results_cityscapes} in terms of ELBO and if available the IWBO (importance weighted bound) \citep{burda2016} with $1000$ samples measured in bits per pixel. Consistent with the language experiments, the traditional dequantization approaches (uniform / variational) are outperformed by Argmax Flows. Interestingly, although argmax flows with softplus thresholding achieves the best ELBO, the argmax flow with Gumbel thresholding approach achieves a better IWBO. The Multinomial Diffusion model performs somewhat worse with 0.37 bpp on test whereas it scored 0.33 bpp on train. Interestingly, this the only model where overfitting was an issue and data augmentation was required, which may explain this portion of the performance difference. For all other models training performance was comparable to test and validation performance. Samples from the different models trained on cityscapes are depicted in Figure~\ref{fig:experiments_cityscapes}. Another interesting point is that coupling flows had difficulty producing coherent text samples (Figure~\ref{fig:samples_text}) but do not suffer from this problem on the cityscapes data which is more image-like. As coupling layers where initially designed for images \citep{dinh2014nice}, they may require adjustments to increase their expressiveness on text.

\section{Social Impact and Conclusion}
\label{sec:conclusion}
\paragraph{Social Impact} The methods described in this paper can be used to learn categorical distributions. For that reason, they can potentially be used to generate high-dimensional categorical data, such as text or image segmentation maps, faster than iterative approaches. Possibly negative influences are the generation of fake media in the form of text, or very unhelpful automated chat bots for customer service. Our work could positively influence new methods for text generation, or improved segmentation for self-driving cars. In addition, our work may also be used for outlier detection to flag fake content. Also, we believe the method in its current form is still distant from direct applications as the ones mentioned above.

\paragraph{Conclusion} In this paper we propose two extensions for Normalizing Flows and Diffusion models to learn categorical data: Argmax Flows and Multinomial Diffusion. Our experiments show that our methods outperform comparable models in terms of negative log-likelihood. In addition, our experiments highlight distinct performance gaps in the field: Between standard ARMs, continuous autoregressive models and non-autoregressive continuous models. This indicates that future work could focus on two sources of decreased performance: 1) when discrete variables are lifted to a continuous space and further 2) when removing autoregressive components.

\textbf{Funding Disclosure}\\
There are no additional sources of funding to disclose, beyond the affiliations of the authors.

\newpage
\bibliography{main.bib}

\providecommand{\latin}[1]{#1}
\makeatletter
\providecommand{\doi}
  {\begingroup\let\do\@makeother\dospecials
  \catcode`\{=1 \catcode`\}=2 \doi@aux}
\providecommand{\doi@aux}[1]{\endgroup\texttt{#1}}
\makeatother
\providecommand*\mcitethebibliography{\thebibliography}
\csname @ifundefined\endcsname{endmcitethebibliography}
  {\let\endmcitethebibliography\endthebibliography}{}
\begin{mcitethebibliography}{36}
\providecommand*\natexlab[1]{#1}
\providecommand*\mciteSetBstSublistMode[1]{}
\providecommand*\mciteSetBstMaxWidthForm[2]{}
\providecommand*\mciteBstWouldAddEndPuncttrue
  {\def\EndOfBibitem{\unskip.}}
\providecommand*\mciteBstWouldAddEndPunctfalse
  {\let\EndOfBibitem\relax}
\providecommand*\mciteSetBstMidEndSepPunct[3]{}
\providecommand*\mciteSetBstSublistLabelBeginEnd[3]{}
\providecommand*\EndOfBibitem{}
\mciteSetBstSublistMode{f}
\mciteSetBstMaxWidthForm{subitem}{(\alph{mcitesubitemcount})}
\mciteSetBstSublistLabelBeginEnd
  {\mcitemaxwidthsubitemform\space}
  {\relax}
  {\relax}

\bibitem[Cooijmans \latin{et~al.}(2017)Cooijmans, Ballas, Laurent,
  G{\"{u}}l{\c{c}}ehre, and Courville]{Cooijmans2017recurrentbatch}
Cooijmans,~T.; Ballas,~N.; Laurent,~C.; G{\"{u}}l{\c{c}}ehre,~{\c{C}}.;
  Courville,~A.~C. Recurrent Batch Normalization. 5th International Conference
  on Learning Representations, {ICLR}. 2017\relax
\mciteBstWouldAddEndPuncttrue
\mciteSetBstMidEndSepPunct{\mcitedefaultmidpunct}
{\mcitedefaultendpunct}{\mcitedefaultseppunct}\relax
\EndOfBibitem
\bibitem[Dai \latin{et~al.}(2019)Dai, Yang, Yang, Carbonell, Le, and
  Salakhutdinov]{dai2019transformerxl}
Dai,~Z.; Yang,~Z.; Yang,~Y.; Carbonell,~J.~G.; Le,~Q.~V.; Salakhutdinov,~R.
  Transformer-XL: Attentive Language Models beyond a Fixed-Length Context.
  Proceedings of the 57th Conference of the Association for Computational
  Linguistics, {ACL} 2019. 2019\relax
\mciteBstWouldAddEndPuncttrue
\mciteSetBstMidEndSepPunct{\mcitedefaultmidpunct}
{\mcitedefaultendpunct}{\mcitedefaultseppunct}\relax
\EndOfBibitem
\bibitem[Uria \latin{et~al.}(2013)Uria, Murray, and Larochelle]{uria2013rnade}
Uria,~B.; Murray,~I.; Larochelle,~H. RNADE: The Real-valued Neural
  Autoregressive Density-estimator. Advances in Neural Information Processing
  Systems. 2013; pp 2175--2183\relax
\mciteBstWouldAddEndPuncttrue
\mciteSetBstMidEndSepPunct{\mcitedefaultmidpunct}
{\mcitedefaultendpunct}{\mcitedefaultseppunct}\relax
\EndOfBibitem
\bibitem[Theis \latin{et~al.}(2016)Theis, van~den Oord, and
  Bethge]{theis2016note}
Theis,~L.; van~den Oord,~A.; Bethge,~M. A note on the evaluation of generative
  models. International Conference on Learning Representations. 2016\relax
\mciteBstWouldAddEndPuncttrue
\mciteSetBstMidEndSepPunct{\mcitedefaultmidpunct}
{\mcitedefaultendpunct}{\mcitedefaultseppunct}\relax
\EndOfBibitem
\bibitem[Ho \latin{et~al.}(2019)Ho, Chen, Srinivas, Duan, and
  Abbeel]{ho2019flow++}
Ho,~J.; Chen,~X.; Srinivas,~A.; Duan,~Y.; Abbeel,~P. Flow++: Improving
  Flow-Based Generative Models with Variational Dequantization and Architecture
  Design. \emph{36th International Conference on Machine Learning}
  \textbf{2019}, \relax
\mciteBstWouldAddEndPunctfalse
\mciteSetBstMidEndSepPunct{\mcitedefaultmidpunct}
{}{\mcitedefaultseppunct}\relax
\EndOfBibitem
\bibitem[Sohl{-}Dickstein \latin{et~al.}(2015)Sohl{-}Dickstein, Weiss,
  Maheswaranathan, and Ganguli]{sohl2015diffusion}
Sohl{-}Dickstein,~J.; Weiss,~E.~A.; Maheswaranathan,~N.; Ganguli,~S. Deep
  Unsupervised Learning using Nonequilibrium Thermodynamics. Proceedings of the
  32nd International Conference on Machine Learning, {ICML}. 2015\relax
\mciteBstWouldAddEndPuncttrue
\mciteSetBstMidEndSepPunct{\mcitedefaultmidpunct}
{\mcitedefaultendpunct}{\mcitedefaultseppunct}\relax
\EndOfBibitem
\bibitem[Ho \latin{et~al.}(2020)Ho, Jain, and Abbeel]{ho2020denoising}
Ho,~J.; Jain,~A.; Abbeel,~P. Denoising Diffusion Probabilistic Models.
  \emph{CoRR} \textbf{2020}, \emph{abs/2006.11239}\relax
\mciteBstWouldAddEndPuncttrue
\mciteSetBstMidEndSepPunct{\mcitedefaultmidpunct}
{\mcitedefaultendpunct}{\mcitedefaultseppunct}\relax
\EndOfBibitem
\bibitem[Song \latin{et~al.}(2020)Song, Meng, and
  Ermon]{song2020denoisingimplicit}
Song,~J.; Meng,~C.; Ermon,~S. Denoising Diffusion Implicit Models. \emph{CoRR}
  \textbf{2020}, \emph{abs/2010.02502}\relax
\mciteBstWouldAddEndPuncttrue
\mciteSetBstMidEndSepPunct{\mcitedefaultmidpunct}
{\mcitedefaultendpunct}{\mcitedefaultseppunct}\relax
\EndOfBibitem
\bibitem[Nichol and Dhariwal(2021)Nichol, and Dhariwal]{nichol2021improved}
Nichol,~A.~Q.; Dhariwal,~P. Improved Denoising Diffusion Probabilistic Models.
  2021; \url{https://openreview.net/forum?id=-NEXDKk8gZ}\relax
\mciteBstWouldAddEndPuncttrue
\mciteSetBstMidEndSepPunct{\mcitedefaultmidpunct}
{\mcitedefaultendpunct}{\mcitedefaultseppunct}\relax
\EndOfBibitem
\bibitem[Rezende and Mohamed(2015)Rezende, and Mohamed]{rezende2015norm}
Rezende,~D.; Mohamed,~S. {Variational Inference with Normalizing Flows}.
  Proceedings of the 32nd International Conference on Machine Learning. 2015;
  pp 1530--1538\relax
\mciteBstWouldAddEndPuncttrue
\mciteSetBstMidEndSepPunct{\mcitedefaultmidpunct}
{\mcitedefaultendpunct}{\mcitedefaultseppunct}\relax
\EndOfBibitem
\bibitem[Nielsen \latin{et~al.}(2020)Nielsen, Jaini, Hoogeboom, Winther, and
  Welling]{nielsen2007survae}
Nielsen,~D.; Jaini,~P.; Hoogeboom,~E.; Winther,~O.; Welling,~M. SurVAE Flows:
  Surjections to Bridge the Gap between VAEs and Flows. \emph{CoRR}
  \textbf{2020}, \emph{abs/2007.02731}\relax
\mciteBstWouldAddEndPuncttrue
\mciteSetBstMidEndSepPunct{\mcitedefaultmidpunct}
{\mcitedefaultendpunct}{\mcitedefaultseppunct}\relax
\EndOfBibitem
\bibitem[Maddison \latin{et~al.}(2014)Maddison, Tarlow, and
  Minka]{Maddison2014Astarsampling}
Maddison,~C.~J.; Tarlow,~D.; Minka,~T. A* Sampling. Advances in Neural
  Information Processing Systems 27: Annual Conference on Neural Information
  Processing Systems. 2014\relax
\mciteBstWouldAddEndPuncttrue
\mciteSetBstMidEndSepPunct{\mcitedefaultmidpunct}
{\mcitedefaultendpunct}{\mcitedefaultseppunct}\relax
\EndOfBibitem
\bibitem[Kool \latin{et~al.}(2019)Kool, Van~Hoof, and Welling]{kool2019}
Kool,~W.; Van~Hoof,~H.; Welling,~M. Stochastic Beams and Where To Find Them:
  The {G}umbel-Top-k Trick for Sampling Sequences Without Replacement.
  Proceedings of the 36th International Conference on Machine Learning.
  2019\relax
\mciteBstWouldAddEndPuncttrue
\mciteSetBstMidEndSepPunct{\mcitedefaultmidpunct}
{\mcitedefaultendpunct}{\mcitedefaultseppunct}\relax
\EndOfBibitem
\bibitem[Winkler \latin{et~al.}(2019)Winkler, Worrall, Hoogeboom, and
  Welling]{winkler2019learning}
Winkler,~C.; Worrall,~D.~E.; Hoogeboom,~E.; Welling,~M. Learning Likelihoods
  with Conditional Normalizing Flows. \emph{CoRR} \textbf{2019},
  \emph{abs/1912.00042}\relax
\mciteBstWouldAddEndPuncttrue
\mciteSetBstMidEndSepPunct{\mcitedefaultmidpunct}
{\mcitedefaultendpunct}{\mcitedefaultseppunct}\relax
\EndOfBibitem
\bibitem[Germain \latin{et~al.}(2015)Germain, Gregor, Murray, and
  Larochelle]{germain2015made}
Germain,~M.; Gregor,~K.; Murray,~I.; Larochelle,~H. Made: Masked autoencoder
  for distribution estimation. International Conference on Machine Learning.
  2015; pp 881--889\relax
\mciteBstWouldAddEndPuncttrue
\mciteSetBstMidEndSepPunct{\mcitedefaultmidpunct}
{\mcitedefaultendpunct}{\mcitedefaultseppunct}\relax
\EndOfBibitem
\bibitem[Kingma and Welling(2014)Kingma, and Welling]{kingma2014auto}
Kingma,~D.~P.; Welling,~M. {Auto-Encoding Variational Bayes}. Proceedings of
  the 2nd International Conference on Learning Representations. 2014\relax
\mciteBstWouldAddEndPuncttrue
\mciteSetBstMidEndSepPunct{\mcitedefaultmidpunct}
{\mcitedefaultendpunct}{\mcitedefaultseppunct}\relax
\EndOfBibitem
\bibitem[Rezende \latin{et~al.}(2014)Rezende, Mohamed, and
  Wierstra]{rezende2014stochasticvariationalinference}
Rezende,~D.~J.; Mohamed,~S.; Wierstra,~D. Stochastic Backpropagation and
  Approximate Inference in Deep Generative Models. Proceedings of the 31th
  International Conference on Machine Learning, {ICML}. 2014\relax
\mciteBstWouldAddEndPuncttrue
\mciteSetBstMidEndSepPunct{\mcitedefaultmidpunct}
{\mcitedefaultendpunct}{\mcitedefaultseppunct}\relax
\EndOfBibitem
\bibitem[Goodfellow \latin{et~al.}(2014)Goodfellow, Pouget-Abadie, Mirza, Xu,
  Warde-Farley, Ozair, Courville, and Bengio]{goodfellow2014generative}
Goodfellow,~I.; Pouget-Abadie,~J.; Mirza,~M.; Xu,~B.; Warde-Farley,~D.;
  Ozair,~S.; Courville,~A.; Bengio,~Y. Generative adversarial nets. Advances in
  neural information processing systems. 2014; pp 2672--2680\relax
\mciteBstWouldAddEndPuncttrue
\mciteSetBstMidEndSepPunct{\mcitedefaultmidpunct}
{\mcitedefaultendpunct}{\mcitedefaultseppunct}\relax
\EndOfBibitem
\bibitem[Dinh \latin{et~al.}(2017)Dinh, Sohl-Dickstein, and
  Bengio]{dinh2016density}
Dinh,~L.; Sohl-Dickstein,~J.; Bengio,~S. {Density estimation using Real NVP}.
  \emph{5th International Conference on Learning Representations, {ICLR}}
  \textbf{2017}, \relax
\mciteBstWouldAddEndPunctfalse
\mciteSetBstMidEndSepPunct{\mcitedefaultmidpunct}
{}{\mcitedefaultseppunct}\relax
\EndOfBibitem
\bibitem[Kingma \latin{et~al.}(2016)Kingma, Salimans, Jozefowicz, Chen,
  Sutskever, and Welling]{kingma2016improved}
Kingma,~D.~P.; Salimans,~T.; Jozefowicz,~R.; Chen,~X.; Sutskever,~I.;
  Welling,~M. {Improved variational inference with inverse autoregressive
  flow}. Advances in Neural Information Processing Systems. 2016; pp
  4743--4751\relax
\mciteBstWouldAddEndPuncttrue
\mciteSetBstMidEndSepPunct{\mcitedefaultmidpunct}
{\mcitedefaultendpunct}{\mcitedefaultseppunct}\relax
\EndOfBibitem
\bibitem[Papamakarios \latin{et~al.}(2017)Papamakarios, Murray, and
  Pavlakou]{papamakarios2017masked}
Papamakarios,~G.; Murray,~I.; Pavlakou,~T. Masked autoregressive flow for
  density estimation. Advances in Neural Information Processing Systems. 2017;
  pp 2338--2347\relax
\mciteBstWouldAddEndPuncttrue
\mciteSetBstMidEndSepPunct{\mcitedefaultmidpunct}
{\mcitedefaultendpunct}{\mcitedefaultseppunct}\relax
\EndOfBibitem
\bibitem[Chen \latin{et~al.}(2018)Chen, Rubanova, Bettencourt, and
  Duvenaud]{chen2018neural}
Chen,~T.~Q.; Rubanova,~Y.; Bettencourt,~J.; Duvenaud,~D.~K. Neural ordinary
  differential equations. Advances in Neural Information Processing Systems.
  2018; pp 6572--6583\relax
\mciteBstWouldAddEndPuncttrue
\mciteSetBstMidEndSepPunct{\mcitedefaultmidpunct}
{\mcitedefaultendpunct}{\mcitedefaultseppunct}\relax
\EndOfBibitem
\bibitem[Song \latin{et~al.}(2019)Song, Meng, and Ermon]{song2019mintnet}
Song,~Y.; Meng,~C.; Ermon,~S. MintNet: Building Invertible Neural Networks with
  Masked Convolutions. Advances in Neural Information Processing Systems 32:
  Annual Conference on Neural Information Processing Systems 2019, NeurIPS
  2019. 2019\relax
\mciteBstWouldAddEndPuncttrue
\mciteSetBstMidEndSepPunct{\mcitedefaultmidpunct}
{\mcitedefaultendpunct}{\mcitedefaultseppunct}\relax
\EndOfBibitem
\bibitem[Perugachi{-}Diaz \latin{et~al.}(2020)Perugachi{-}Diaz, Tomczak, and
  Bhulai]{perugachi2020idensenets}
Perugachi{-}Diaz,~Y.; Tomczak,~J.~M.; Bhulai,~S. Invertible DenseNets.
  \emph{CoRR} \textbf{2020}, \emph{abs/2010.02125}\relax
\mciteBstWouldAddEndPuncttrue
\mciteSetBstMidEndSepPunct{\mcitedefaultmidpunct}
{\mcitedefaultendpunct}{\mcitedefaultseppunct}\relax
\EndOfBibitem
\bibitem[Nielsen and Winther(2020)Nielsen, and
  Winther]{nielsen2020closingdequantizationgap}
Nielsen,~D.; Winther,~O. Closing the Dequantization Gap: PixelCNN as a
  Single-Layer Flow. Advances in Neural Information Processing Systems 33:
  Annual Conference on Neural Information Processing Systems 2020, {NeurIPS}.
  2020\relax
\mciteBstWouldAddEndPuncttrue
\mciteSetBstMidEndSepPunct{\mcitedefaultmidpunct}
{\mcitedefaultendpunct}{\mcitedefaultseppunct}\relax
\EndOfBibitem
\bibitem[Tran \latin{et~al.}(2019)Tran, Vafa, Agrawal, Dinh, and
  Poole]{discrete2019tran}
Tran,~D.; Vafa,~K.; Agrawal,~K.; Dinh,~L.; Poole,~B. Discrete Flows: Invertible
  Generative Models of Discrete Data. \emph{ICLR 2019 Workshop DeepGenStruct}
  \textbf{2019}, \relax
\mciteBstWouldAddEndPunctfalse
\mciteSetBstMidEndSepPunct{\mcitedefaultmidpunct}
{}{\mcitedefaultseppunct}\relax
\EndOfBibitem
\bibitem[Hoogeboom \latin{et~al.}(2019)Hoogeboom, Peters, van~den Berg, and
  Welling]{hoogeboom2019integer}
Hoogeboom,~E.; Peters,~J. W.~T.; van~den Berg,~R.; Welling,~M. Integer Discrete
  Flows and Lossless Compression. Neural Information Processing Systems 2019,
  NeurIPS 2019. 2019; pp 12134--12144\relax
\mciteBstWouldAddEndPuncttrue
\mciteSetBstMidEndSepPunct{\mcitedefaultmidpunct}
{\mcitedefaultendpunct}{\mcitedefaultseppunct}\relax
\EndOfBibitem
\bibitem[van~den Berg \latin{et~al.}(2020)van~den Berg, Gritsenko, Dehghani,
  S{\o}nderby, and Salimans]{berg2020idfpp}
van~den Berg,~R.; Gritsenko,~A.~A.; Dehghani,~M.; S{\o}nderby,~C.~K.;
  Salimans,~T. {IDF++:} Analyzing and Improving Integer Discrete Flows for
  Lossless Compression. \emph{CoRR} \textbf{2020}, \emph{abs/2006.12459}\relax
\mciteBstWouldAddEndPuncttrue
\mciteSetBstMidEndSepPunct{\mcitedefaultmidpunct}
{\mcitedefaultendpunct}{\mcitedefaultseppunct}\relax
\EndOfBibitem
\bibitem[Ziegler and Rush(2019)Ziegler, and Rush]{ziegler2019latent}
Ziegler,~Z.~M.; Rush,~A.~M. Latent Normalizing Flows for Discrete Sequences.
  Proceedings of the 36th International Conference on Machine Learning, {ICML}.
  2019\relax
\mciteBstWouldAddEndPuncttrue
\mciteSetBstMidEndSepPunct{\mcitedefaultmidpunct}
{\mcitedefaultendpunct}{\mcitedefaultseppunct}\relax
\EndOfBibitem
\bibitem[Lippe and Gavves(2020)Lippe, and Gavves]{lippe2020categorical}
Lippe,~P.; Gavves,~E. Categorical Normalizing Flows via Continuous
  Transformations. \emph{CoRR} \textbf{2020}, \emph{abs/2006.09790}\relax
\mciteBstWouldAddEndPuncttrue
\mciteSetBstMidEndSepPunct{\mcitedefaultmidpunct}
{\mcitedefaultendpunct}{\mcitedefaultseppunct}\relax
\EndOfBibitem
\bibitem[Song \latin{et~al.}(2020)Song, Sohl{-}Dickstein, Kingma, Kumar, Ermon,
  and Poole]{song2020scorebasedSDEs}
Song,~Y.; Sohl{-}Dickstein,~J.; Kingma,~D.~P.; Kumar,~A.; Ermon,~S.; Poole,~B.
  Score-Based Generative Modeling through Stochastic Differential Equations.
  \emph{CoRR} \textbf{2020}, \emph{abs/2011.13456}\relax
\mciteBstWouldAddEndPuncttrue
\mciteSetBstMidEndSepPunct{\mcitedefaultmidpunct}
{\mcitedefaultendpunct}{\mcitedefaultseppunct}\relax
\EndOfBibitem
\bibitem[Al{-}Rfou \latin{et~al.}(2019)Al{-}Rfou, Choe, Constant, Guo, and
  Jones]{alrfou2019transformer}
Al{-}Rfou,~R.; Choe,~D.; Constant,~N.; Guo,~M.; Jones,~L. Character-Level
  Language Modeling with Deeper Self-Attention. The Thirty-Third {AAAI}
  Conference on Artificial Intelligence, {AAAI} 2019. 2019\relax
\mciteBstWouldAddEndPuncttrue
\mciteSetBstMidEndSepPunct{\mcitedefaultmidpunct}
{\mcitedefaultendpunct}{\mcitedefaultseppunct}\relax
\EndOfBibitem
\bibitem[Huang \latin{et~al.}(2017)Huang, Liu, Van Der~Maaten, and
  Weinberger]{huang2017densely}
Huang,~G.; Liu,~Z.; Van Der~Maaten,~L.; Weinberger,~K.~Q. Densely connected
  convolutional networks. Proceedings of the IEEE conference on computer vision
  and pattern recognition. 2017; pp 4700--4708\relax
\mciteBstWouldAddEndPuncttrue
\mciteSetBstMidEndSepPunct{\mcitedefaultmidpunct}
{\mcitedefaultendpunct}{\mcitedefaultseppunct}\relax
\EndOfBibitem
\bibitem[Burda \latin{et~al.}(2016)Burda, Grosse, and Salakhutdinov]{burda2016}
Burda,~Y.; Grosse,~R.~B.; Salakhutdinov,~R. Importance Weighted Autoencoders.
  4th International Conference on Learning Representations. 2016\relax
\mciteBstWouldAddEndPuncttrue
\mciteSetBstMidEndSepPunct{\mcitedefaultmidpunct}
{\mcitedefaultendpunct}{\mcitedefaultseppunct}\relax
\EndOfBibitem
\bibitem[Dinh \latin{et~al.}(2015)Dinh, Krueger, and Bengio]{dinh2014nice}
Dinh,~L.; Krueger,~D.; Bengio,~Y. {NICE: Non-linear independent components
  estimation}. \emph{3rd International Conference on Learning Representations,
  {ICLR}, Workshop Track Proceedings} \textbf{2015}, \relax
\mciteBstWouldAddEndPunctfalse
\mciteSetBstMidEndSepPunct{\mcitedefaultmidpunct}
{}{\mcitedefaultseppunct}\relax
\EndOfBibitem
\bibitem[Cordts \latin{et~al.}(2016)Cordts, Omran, Ramos, Rehfeld, Enzweiler,
  Benenson, Franke, Roth, and Schiele]{Cordts2016cityscapes}
Cordts,~M.; Omran,~M.; Ramos,~S.; Rehfeld,~T.; Enzweiler,~M.; Benenson,~R.;
  Franke,~U.; Roth,~S.; Schiele,~B. The Cityscapes Dataset for Semantic Urban
  Scene Understanding. 2016 {IEEE} Conference on Computer Vision and Pattern
  Recognition, {CVPR}. 2016; pp 3213--3223\relax
\mciteBstWouldAddEndPuncttrue
\mciteSetBstMidEndSepPunct{\mcitedefaultmidpunct}
{\mcitedefaultendpunct}{\mcitedefaultseppunct}\relax
\EndOfBibitem
\end{mcitethebibliography}
\bibliographystyle{achemso}

% \newpage
% \input{sections/checklist}

\newpage
\appendix

\section{Numerically stable Multinomial Diffusion in log space}
\label{appendix:numerically_stable_diffusion}
In this section we explain how Multinomial Diffusion models can be implemented in a numerically safe manner in log-space. Note that in addition to this appendix with pseudo-code, the actual source code will also be released. First we define a few helper functions:
\begin{lstlisting}
def log_add_exp(a, b):
    maximum = max(a, b)
    return maximum + log(exp(a - maximum) + exp(b - maximum))
    
def log_sum_exp(x):
    maximum = max(x, dim=1, keepdim=True)
    return maximum + log(exp(x - maximum).sum(dim=1))

def index_to_log_onehot(x, num_classes):
    # Assume that onehot axis is inserted at dimension 1
    x_onehot = one_hot(x, num_classes)
    
    # Compute in log-space, extreme low values are later
    # filtered out by log sum exp calls.
    log_x = log(x_onehot.clamp(min=1e-40))
    return log_x

def log_onehot_to_index(log_x):
    return log_x.argmax(1)
    
def log_1_min_a(a):
    return log(1 - a.exp() + 1e-40)
\end{lstlisting}

Then we can initialize the variables we are planning to utilize for the multinomial diffusion model. This is done with float64 variables to limit the precision loss in the \texttt{log\_1\_min\_a} computation. Since these are precomputed and later converted to float32, there is no meaningful increase in computation time. 
\begin{lstlisting}
alphas = init_alphas() 
log_alpha = np.log(alphas)
log_cumprod_alpha = np.cumsum(log_alpha)

log_1_min_alpha = log_1_min_a(log_alpha)
log_1_min_cumprod_alpha = log_1_min_a(log_cumprod_alpha)
\end{lstlisting}

Then we can define the functions that we utilize to compute the log probabilities of the categorical distributions of the forward process. The functions below compute the probability vectors for $q(\vx_t | \vx_{t-1})$, $q(\vx_t | \vx_{0})$ and $q(\vx_{t-1} | \vx_t, \vx_{0})$.
\begin{lstlisting}
def q_pred_one_timestep(log_x_t, t):
    # Computing alpha_t * E[xt] + (1 - alpha_t) 1 / K
    log_probs = log_add_exp(
        log_x_t + log_alpha[t],
        log_1_min_alpha[t] - log(num_classes)
    )
    return log_probs

def q_pred(log_x0, t):
    log_probs = log_add_exp(
        log_x0 + log_cumprod_alpha[t],
        log_1_min_cumprod_alpha[t] - log(num_classes)
    )
    return log_probs

def q_posterior(log_x0, log_x_t, t):
    # Kronecker delta peak for q(x0 | x1, x0).
    if t == 0:
        log_probs_xtmin = log_x0
    else:
        log_probs_xtmin = q_pred(log_x0, t - 1)
    
    # Note log_x_t is used not x_tmin, subtle and not straightforward
    # why this is true. Corresponds to Algorithm 1.
    unnormed_logprobs = log_probs_xtmin + q_pred_one_timestep(log_x_t, t) 
        
    log_probs_posterior = unnormed_logprobs - log_sum_exp(unnormed_logprobs)
    return log_probs_posterior
\end{lstlisting}

Some magic is happening in \texttt{q\_pred\_one\_timestep}. Recall that at some point we need to compute $\mathcal{C}(\vx_t | (1 - \beta_t) \vx_{t-1} + \beta_t / K )$ for different values of $\vx_t$, which when treated as a function outputs $(1 - \beta_t) + \beta_t / K$ if $\vx_t = \vx_{t-1}$ and $\beta_t / K$ otherwise. This function is symmetric, meaning that $\mathcal{C}(\vx_t | (1 - \beta_t) \vx_{t-1} + \beta_t / K ) = \mathcal{C}(\vx_{t-1} | (1 - \beta_t) \vx_{t} + \beta_t / K )$. This is why we can switch the conditioning and immediately return the different probability vectors for $\vx_t$. This also corresponds to Equation~\ref{eq:q_posterior}.

Then using the \texttt{q\_posterior} function as parametrization we predict the probability vector for $p(\vx_{t-1} | \vx_t)$ using a neural network.
\begin{lstlisting}
def p_pred(log_x_t, t):
    x_t = log_onehot_to_index(log_x_t)
    log_x_recon = logsoftmax(neuralnet(x_t, t))
    log_model_pred = q_posterior(log_x_recon, log_x_t, t)
    return log_model_pred
\end{lstlisting}

And then finally we can compute the loss term $L_t$ using the KL divergence for categorical distributions:
\begin{lstlisting}
def categorical_kl(log_prob_a, log_prob_b):
    kl = (log_prob_a.exp() * (log_prob_a - log_prob_b)).sum(dim=1)
    return kl

def compute_Lt(log_x0, log_x_t, t):
    log_true_prob = q_posterior(log_x0, log_x_t, t)
    log_model_prob = p_pred(log_x_t, t)
    kl = categorical_kl(log_true_prob, log_model_prob)
    loss = sum_except_batch(kl)
    return loss
\end{lstlisting}

Coincidentally this code even works for $L_0$ because $\vx_0$ is onehot and then: 
$$-\log \mathcal{C}(\vx_0 | \hat{\vx}_{0}) - \sum_k \vx_{0,k} \log  \hat{\vx}_{0,k} = \sum_k \vx_{0,k} [\underbrace{\log \vx_{0,k}}_{0 \text{ or } \log 0} - \log  \hat{\vx}_{0,k}] = \mathrm{KL}(\mathcal{C}(\vx_{0})||\mathcal{C}(\hat{\vx}_{0})),$$ where in the last term $\vx_{0}$ and $\hat{\vx}_{0}$ are probability vectors and $0 \log 0$ is defined to be $0$.

\newpage
\section{Experimental details}
\label{app:experimental_details}
This section gives details on experimental setup, architectures and optimization hyperparameters. In addition, the code to reproduce experiments will be released publicly.

\paragraph{Diffusion settings}
For diffusion we use the cosine schedule for $\{\alpha_t\}$ from \cite{nichol2021improved} with the difference that what was previously $\sqrt{\bar{\alpha}_t}$ is now $\bar{\alpha}_t$, so that their factor $\sqrt{\bar{\alpha}_t}$ for the Gaussian mean is equal to our factor $\bar{\alpha}_t$ for categorical parameters. Specifically, our $\bar{\alpha}_t$ are defined using:
\begin{equation*}
    \bar{\alpha}_t =\frac{f(t)}{f(0)} \quad f(t) = \cos \left( \frac{t/T + s}{1  + s} \cdot \frac{\pi}{2} \right), \quad s = 0.008,
\end{equation*}
where $T$ is the total number of diffusion steps. \citet{nichol2021improved} show that instead of sampling $t$ uniformly, variance is reduced when $t$ is importance-sampled with $q(t) \propto \sqrt{\mathbb{E}[L_t^2]}$, which is estimated using training statistics, and we use their approach. The objective can be summarized as:
\small
\begin{equation}
    \log P(\vx_0) \geq \mathbb{E}_{t \sim q(t), \vx_t \sim q(\vx_t | \vx_0)} 
    \left[ - \frac{1}{q(t)} \mathrm{KL} \big{(}q(\vx_{t-1} | \vx_t, \vx_0) | p(\vx_{t-1} | \vx_t) \right].
\end{equation}
\normalsize

\paragraph{Gumbel properties}
In Table~\ref{tab:gumbel} a useful overview of Gumbel properties are given. These equations can be used to sample and compute the likelihood of the (truncated) Gumbel distributions. For a more extensive treatment see \citep{Maddison2014Astarsampling,kool2019}.
\begin{table}[H]
    \centering
    \caption{Summary of Gumbel properties.}
    \label{tab:gumbel}
    \scalebox{.67}{
    \begin{tabular}{l p{3.95cm} p{4.1cm}}
    \toprule
    Description & $\log p$ & Sample \\ \midrule
    \multirow{2}{*}{$\mathrm{Gumbel}(g | \phi)$} & \multirow{2}{*}{$\phi - g - \exp(\phi - g)$}  & $g = -\log (-\log(u)) + \phi$ \newline $u \sim \mathcal{U}(0, 1)$
    \\ \midrule
    \multirow{2}{*}{$\max_i \mathrm{Gumbel}(g_i | \phi)$} & $\log \mathrm{Gumbel}(g_{\max} | \phi_{\max})$ \newline $\phi_{\max} = \log \sum_i \exp \phi_i$  & $g_{\max} \sim \mathrm{Gumbel}(\phi_{\max})$ \newline $\phi_{\max} = \log \sum_i \exp \phi_i$
    \\ \midrule
    \multirow{2}{*}{\small$\mathrm{TruncGumbel}(g| \phi, T)$} & \small $\phi - g - \exp(\phi - g) + \exp(\phi - T)$ \newline \text{ if } $g < T$ \text{ else } $-\infty$   & \small $g = \phi - \log(\exp(\phi - T) - \log u)$ \newline $u \sim \mathcal{U}(0, 1)$ \\ \bottomrule
    \end{tabular}}
\end{table}

\subsection{Language Modelling}
For the language modelling experiments we utilize the standard \texttt{text8} dataset with sequence length $256$ and \texttt{enwik8} dataset with sequence length $320$. The train/val/test splits are 90000000/5000000/5000000 for both \texttt{text8} and \texttt{enwik8}, as is standard in literature. The Multinomial Text Diffusion models are trained for $300$ epochs, whereas the Argmax Flows are trained for $40$ epochs, with the exception of the Argmax Coupling Flow on enwik8 which only needs to be trained for $20$ epochs. Further details are presented in Tables~\ref{tab:details_text_models} and \ref{tab:architecture_text_models}. In addition, the code to reproduce results will be publicly available. There are no known ethics issues with these datasets at the time of writing.

\begin{table}[H]
    \centering
    \caption{Optimization details for text models.}
    \label{tab:details_text_models}
    \scalebox{.85}{
    \begin{tabular}{l l l l l l l l l  }
    \toprule
    Model & batch size & lr & lr decay & optimizer &  dropout \\ \midrule
    Multinomial Text Diffusion (text8) &  32 & 0.0001 & 0.99 & Adam & 0  \\
    Multinomial Text Diffusion (enwik8) &  32 & 0.0001 & 0.99 & Adam & 0  \\
    Argmax AR Flow (text8) &  64 & 0.001 & 0.995 & Adam & 0.25  \\
    Argmax AR Flow (enwik8) &  64 & 0.001 & 0.995 & Adam & 0.25  \\
    Argmax Coupling Flow (text8) & 16 & 0.001 & 0.995 & Adamax & 0.05  \\
    Argmax Coupling Flow (enwik8) & 32 & 0.001 & 0.995 & Adamax & 0.1  \\ \bottomrule
    % Discrete Flow & - &  \\
    % VAE & 5.04 & 4.84 \\
    \end{tabular}}
\end{table}

\begin{table}[H]
    \centering
    \caption{Architecture description for text models.}
    \label{tab:architecture_text_models}
    \scalebox{.85}{
    \begin{tabular}{l l l l l l l l l  }
    \toprule
    Model & Architecture description \\ \midrule
    Multinomial Text Diffusion (text8) & 12-layer transformer 8 global, 8 local heads / 1000 diffusion steps  \\
    Multinomial Text Diffusion (enwik8) & 12-layer transformer 8 global, 8 local heads / 4000 diffusion steps  \\
    Argmax AR Flow (text8) & 2-layer LSTM, 2048 hidden units  \\
    Argmax AR Flow (enwik8) & 2-layer LSTM, 2048 hidden units \\
    Argmax Coupling Flow (text8) & 2-layer bi-directional LSTM, 512 hidden units \\
    Argmax Coupling Flow (enwik8) & 2-layer bi-directional LSTM, 768 hidden units  \\ \bottomrule
    % Discrete Flow & - &  \\
    % VAE & 5.04 & 4.84 \\
    \end{tabular}}
\end{table}

\subsection{Cityscapes}

\paragraph{Preprocessing}
The Cityscapes \citep{Cordts2016cityscapes} segmentation maps are re-sampled to a $32$ by $64$ pixel image using nearest neighbour interpolation. The original segmentation maps are downloaded from \url{https://www.cityscapes-dataset.com/downloads/} where all files are contained in \texttt{gtFine\_trainvaltest.zip}. Note that we train on a $8$-class problem since we only consider what is called the \texttt{category\_id} field in torchvision. We re-purpose the validation set as test set, containing $500$ maps. The original train set containing $2975$ maps is split into $2500$ maps for training and $475$ maps for validation. The original test set is not utilized. To aid reproducibility we will publish source code that includes the preprocessing and the dataloaders. There are no known ethics issues with the segmentation maps at the time of writing. License is located at \url{https://www.cityscapes-dataset.com/license/}.

\paragraph{Architectures} For Cityscapes all models utilize the same architectures, although they represent a different part for their respective model designs. The density model $p(\vv)$ consist of $4$ levels with $10$ subflows each, separated by squeeze layers, where each subflow consists of a $1$ $\times$ $1$ convolution and an affine coupling layer. The coupling layers are parametrized by DenseNets \citep{huang2017densely}. The same model is used for the latent distribution in the VAE (usually referred to as $p(\vz)$ in literature). The probabilistic inverse $q(\vv | \vx)$ is modelled by a single level flow that has $8$ subflows, again consisting of affine coupling layers and $1$ $\times$ $1$ convolutions. To condition on $\vx$ it is processed by a DenseNet which outputs a representation for the coupling layers that is concatenated to the original input. The same model is utilized to parametrize the VAE encoder (commonly referred to as $q(\vz | \vx)$). The VAE additionally has a model for the decoder $p(\vx | \vz)$ which is parametrized by a DenseNet which outputs the parameters for a categorical distribution. The models are optimized using the same settings, and no hyperparameter search was performed. Specifically, the models are optimized with minibatch size $64$ for $2000$ epochs with the Adamax optimizer with learning rate $0.001$ and a linear learning rate warmup of $10$ epochs and a decay factor of $0.995$.

\subsection{Range of considered hyperparameters}
For Multinomial Text Diffusion we experimented with the depth of transformers $\{1, 2, 4, 8, 12, 16, 20\}$ and the hidden size $\{128, 256, 512, 1024\}$. We found that models with depth $12$ and $512$ could be trained in a reasonable amount of time while giving good performance. For the cityscapes experiments no hyperparameter search was performed.

\subsection{Details on latent normalizing flows for text8}
We utilize the official code repository from \cite{ziegler2019latent} in here\footnote{\url{https://github.com/harvardnlp/TextFlow}}. The original code utilizes $10$ ELBO samples, which is relatively expensive. For that reason we instead opt for $1$ ELBO sample and find it gives similar results. The batch size is increased from $16$ to $32$. Additionally we reduce the KL scheduling from $4$ initial $10^{-5}$ epochs to only $2$ initial $10^{-5}$ epoch and we anneal linearly over the next $4$ epochs instead of over the next $10$ epochs. In total the models are optimized for $30$ epochs. We verify that the resulting models still achieve similar performance on the Penn Tree Bank experiment compared to the original paper in terms of ELBO values: Our hyperparameter setup for AF/AF achieves slightly better performance with 1.46 versus 1.47 bpc and for IAF/SCF achieves slightly worse 1.78 versus 1.76 bpc.

\subsection{Computing infrastructure}
Experiments where run on NVIDIA-GTX 1080Ti GPUs, CUDA 10.1 with Python version 3.7.6 in Pytorch 1.5.1 or 1.7.1. 

\section{Reproducing Discrete Flows}
\label{sec:reproducing_discrete_flows}
\begin{figure}[h]
    \centering
    \begin{subfigure}{.4\textwidth}
    \centering
    \includegraphics[width=\textwidth]{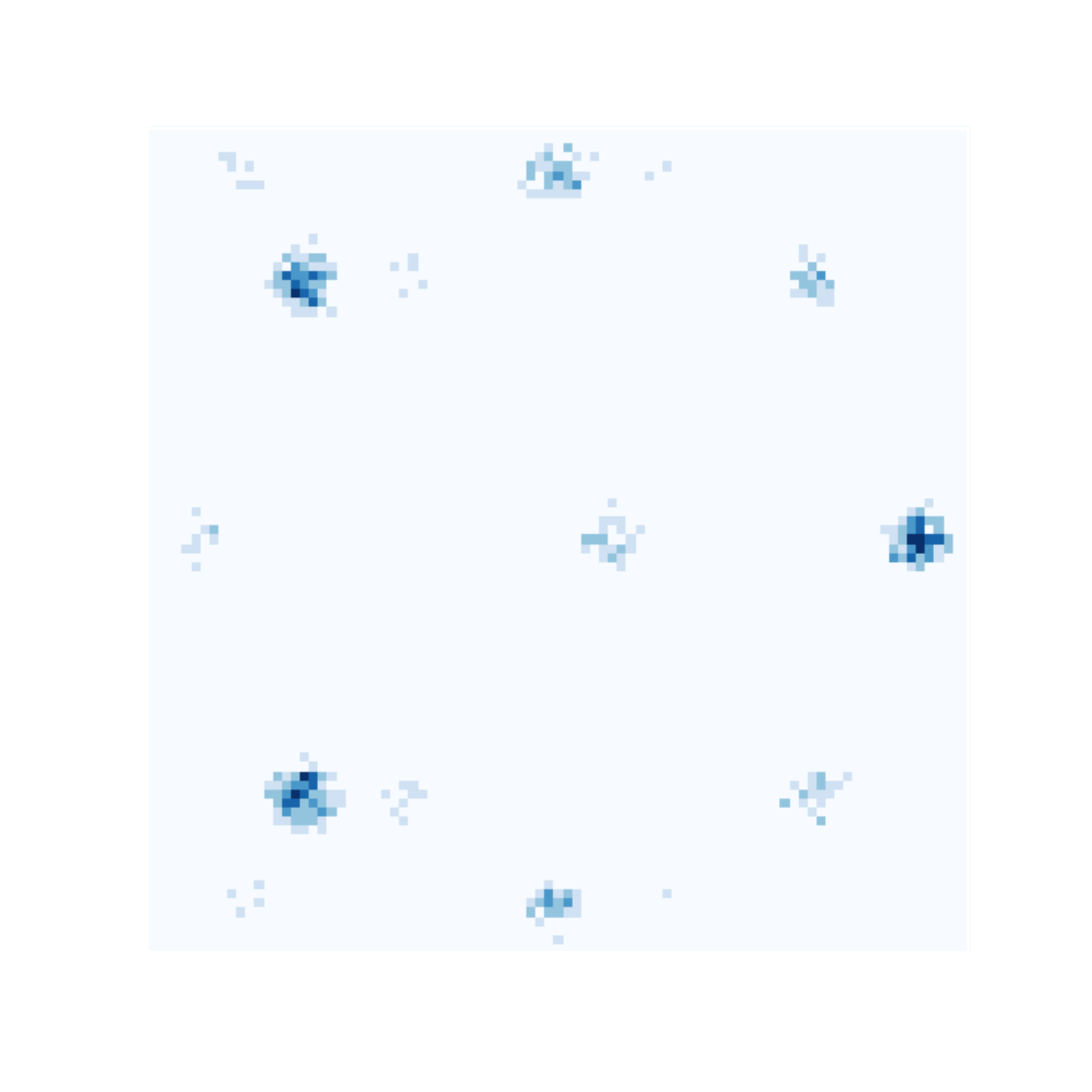}
    \caption{Samples from Discrete Flow using a single layer, taken from \citep{discrete2019tran}.}
    \end{subfigure}
    \hspace{1cm}
    \begin{subfigure}{.4\textwidth}
    \includegraphics[width=\textwidth]{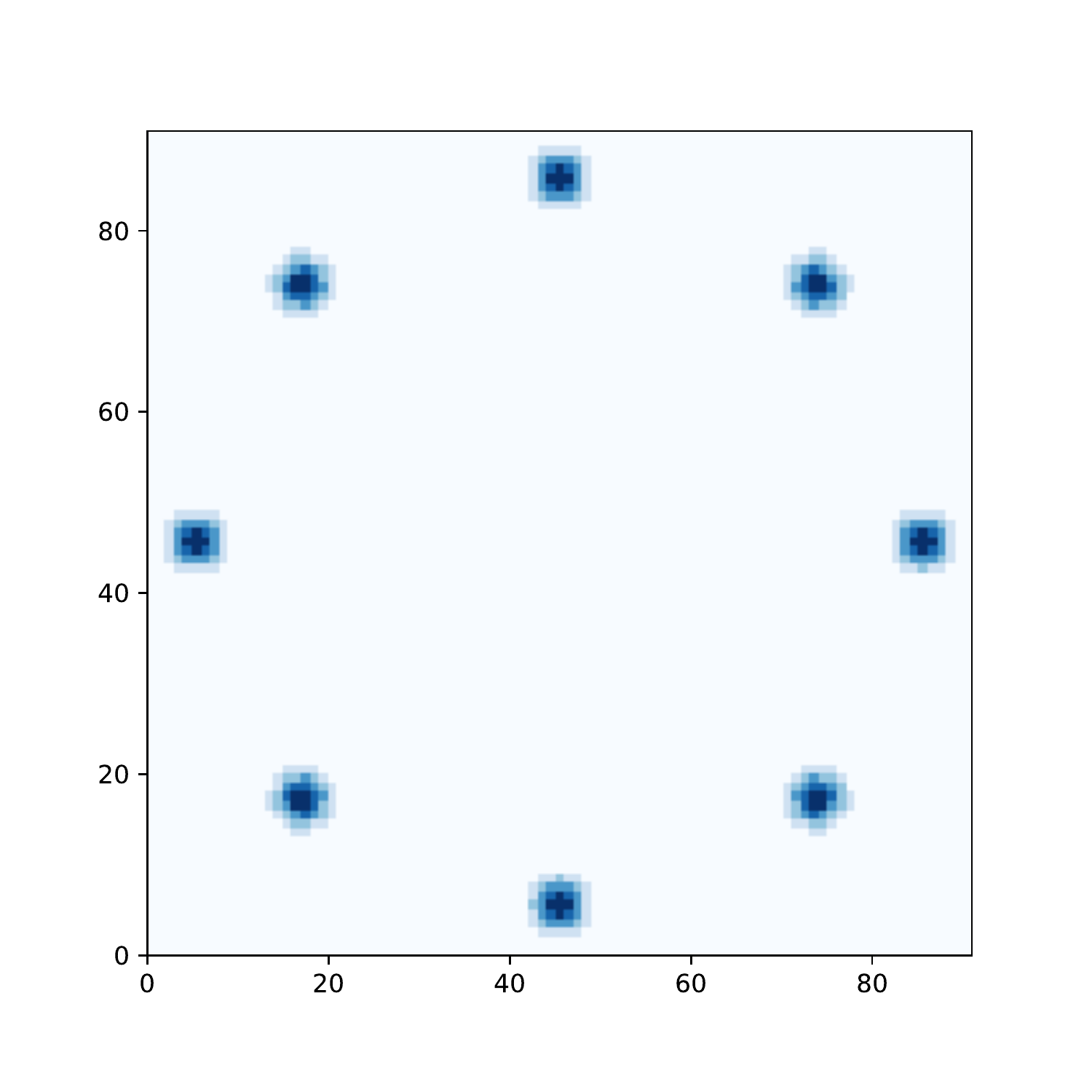}
    \caption{Samples from the quantized 8 Gaussians data distribution.}
    \end{subfigure}
    \begin{subfigure}{.4\textwidth}
    \includegraphics[width=\textwidth]{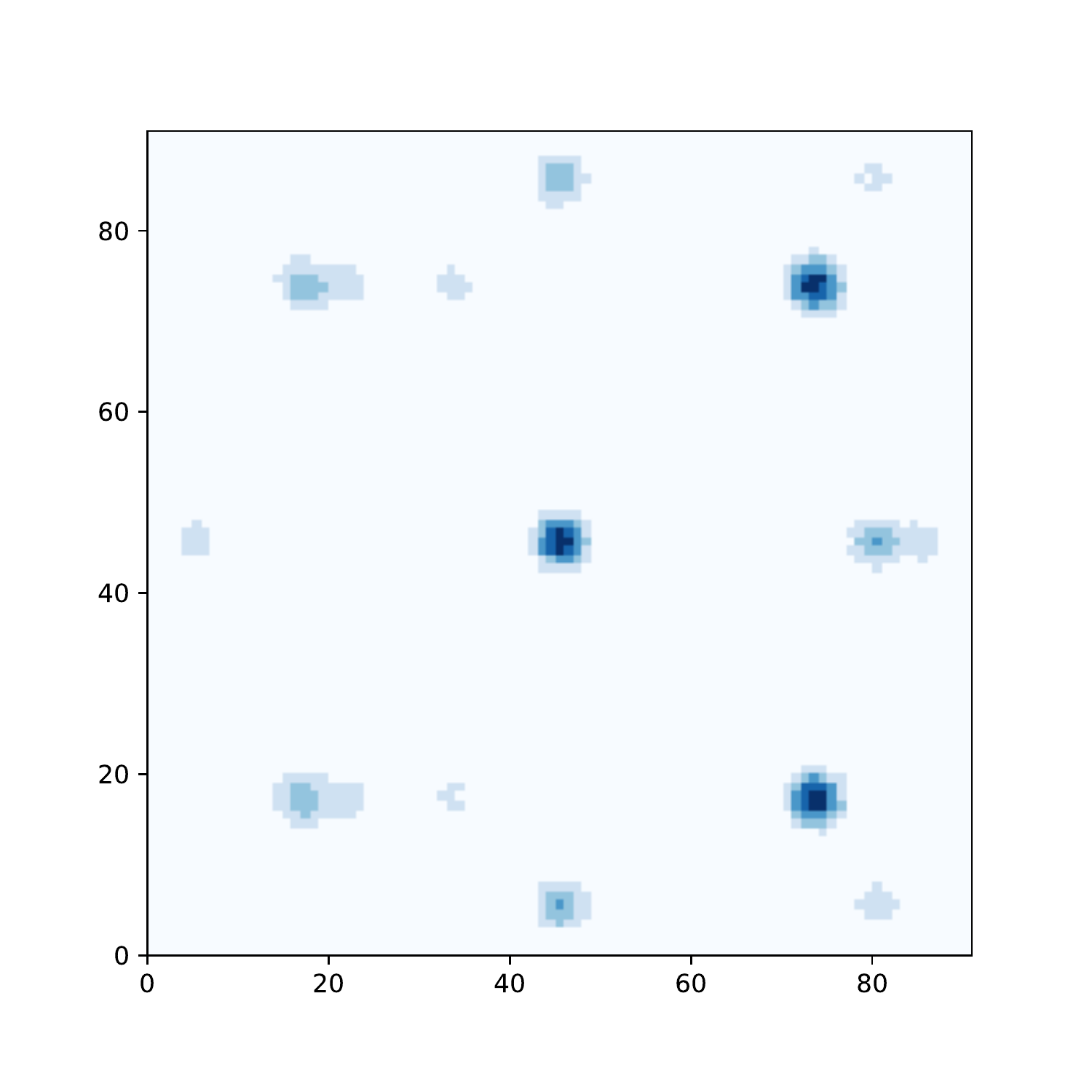}
    \caption{Samples from the Discrete Flows PyTorch re-implementation, achieving 7.0 nats.}
    \end{subfigure}
    \hspace{1cm}
    \begin{subfigure}{.4\textwidth}
    \includegraphics[width=\textwidth]{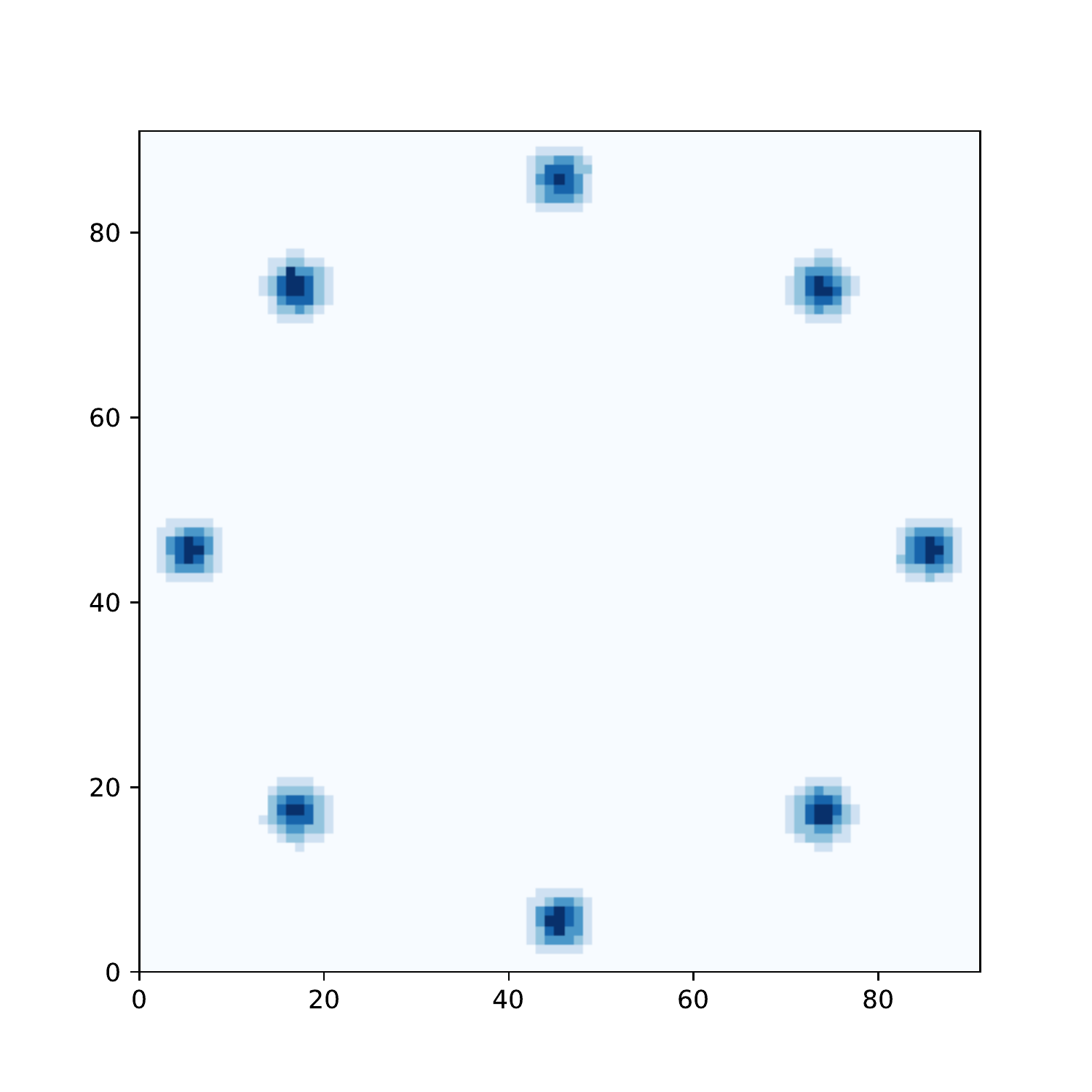}
    \caption{Probability mass of our Argmax Flow using a single layer, achieving 6.32 nats.}
    \end{subfigure}
    \caption{Reproduction of the quantized eight Gaussians experiment. Plots show either the probability mass function or weighted number of samples (which will tend towards the pmf).}
    \label{fig:8gaussiansexp}
\end{figure}

In this section we detail our efforts to reproduce the results from discrete flows \citep{discrete2019tran}. Specifically, we are interested in the discrete flows models that map to \textit{factorized} distributions, for instance the discrete bipartite (coupling) flow. We avoid situations where an autoregressive base distribution is used, it may be difficult to identify how much the flow is actually learning versus the ARM as base. For this paper an official implementation was released at \url{https://github.com/google/edward2/blob/master/edward2/tensorflow/layers/} in the files \texttt{discrete\_flows.py} and \texttt{utils.py}. However, this codebase contains only the high-level modules and code for the toy example, it does not contain the specific code related to the language experiments. These high-level modules and the toy problem were ported to PyTorch here: \url{https://github.com/TrentBrick/PyTorchDiscreteFlows}. Using this codebase, we were able to compare on the quantized eight Gaussians toy dataset, as depicted in Figure~\ref{fig:8gaussiansexp}. In this experiment we clearly see that argmax flows outperform discrete flows both numerically (6.32 versus 7.0 nats) and visually by comparing the samples or probability mass function. 

Subsequent efforts by others to reproduce the language experiments failed (see \url{https://github.com/TrentBrick/PyTorchDiscreteFlows/issues/1}). In another work, \citet{lippe2020categorical} also noticed the difficulty of getting discrete flows to succesfully optimize, as detailed in the set shuffling/summation experiment corresponding to Table 5 in the paper.

For this paper we also tried to reproduce the language experiments. After verifying the correctness of the \texttt{one\_hot\_argmax}, \texttt{one\_hot\_minus} and \texttt{one\_hot\_add} functions in \url{https://github.com/TrentBrick/PyTorchDiscreteFlows}, we implemented an autoregressive discrete flow layer with an expressive network, in an effort to limit the accumulated gradient bias. Recall that an autoregressive layer is more expressive than a coupling layer as it has more dependencies between dimensions. As can be seen in Table~\ref{tab:discrete_flows_reimplementation} our re-implementation also performed considerably worse, matching the experience of the others described above.
\begin{table}[H]
    \centering
    \caption{Discrete Flows on \texttt{text8}. Note that AR is more expressive than coupling.}
    \label{tab:discrete_flows_reimplementation}
    \scalebox{.95}{
    \begin{tabular}{l c }
    \toprule
    Model & text8 (bpc) \\ \midrule
    Discrete Flows from paper (coupling, factorized base, without scale) & 1.29 \\
    Discrete Flows from paper (coupling, factorized base, with scale) & 1.23 \\
    Discrete Flows reimplementation (AR, factorized base, without scale) & 4.13 \\ \midrule
    Argmax Flow, AR (ours) & 1.38 \\
    Argmax Coupling Flow (ours) & 1.80 \\
    \bottomrule
    \end{tabular}}
\end{table}

\paragraph{Final remarks}
We have had extensive contact with the authors of \citep{discrete2019tran} to resolve this issue over the course of several months. Unfortunately it is not possible for them to share the code for the language flows due to internal dependencies. Also, we have not been able to find any implementation of discrete flows online that achieves the reported performance on text. The authors generously offered to look at our reimplementation, which we have shared with them. At the time of writing we have not yet heard anything back on the code. For the reasons described in this appendix, we currently assume that the language experiments in discrete flows are not reproducible.

\newpage

\section{Additional experiments}
\label{sec:additional_experiments}
% \begin{table}
%     \centering
%     \caption{Example of the trade-offs for Cartesian products of Argmax Flows, in a hypothetical problem with $K = 2000$ classes.}
%     \label{tab:hypothetical_cartesian}
%     \scalebox{.8}{
%     \begin{tabular}{r r r r r r}
%     \toprule
%     $M$ & $d_m$ & neighbours & max distance & latent dimensions \\ \midrule
%     % Discrete Flow & - &  \\
%     % VAE & 5.04 & 4.84 \\ 
%     $2000$ & $1$ & $1999$ & $1$ & $2000$ \\
%     $45$ & $2$ & $45$ & $2$ & $90$ \\ 
%     $13$ & $3$ & $13$ & $3$ & $39$ \\ 
%     $2$ & $11$ & $2$ & $11$ & $22$  \\\bottomrule
%     \end{tabular}}
% \end{table}

A comparison of the performance for Cartesian products with different bases is shown in Table \ref{tab:cartesian_products}. Note that this experiment was performed using a somewhat smaller architecture then in the main text. As can be seen, the performance difference between different Cartesian products is relatively small. The performance does decreases slightly over larger base numbers, indicating that it is better to choose a small base that results in fewer overall dimensions.
\begin{table}[H]
    \centering
    \caption{Cartesian Products with different base numbers trained using a slightly smaller version of the Argmax AR Flow on \texttt{text8}.}
    \label{tab:cartesian_products}
    \scalebox{.95}{
    \begin{tabular}{l c }
    \toprule
    Model & text8 (bpc) \\ \midrule
    $d_m=1, M=27$ & 1.45 \\ 
    $d_m=2, M=6$ & 1.44 \\
    $d_m=3, M=3$ & 1.44 \\
    $d_m=5, M=2$ & 1.44 \\
    \bottomrule
    \end{tabular}}
    \vspace{-.5cm}
\end{table}

A comparison of sampling time speeds are shown in Table~\ref{tab:time_experiment}. A couple of orders in magnitude difference can be seen comparing autoregressive versus non-autoregressive models. This highlights the importance of researching generative models that can be built from non-autoregressive components. The main source of difference between our coupling approach and IAF/SCF is that we utilize mixture of discretized logistics \citep{ho2019flow++} as coupling transformation, which requires a iterative process to invert over 1 dimension. The multinomial diffusion takes in-between the time of autoregressive and coupling models. Also reducing steps reduces the required sampling time, as is expected.
\begin{table}[H]
    \centering
    \vspace{-.25cm}
    \caption{Comparison of different methods in terms of sample time.  Sample time is measured by generating a single text sample of length 256 averaged over $10$ runs, unless specified otherwise. }
    \label{tab:time_experiment}
    \begin{tabular}{l l l c c c}
    \toprule
    Model type & & Model &  Sample time (s) \\ \midrule
    \multirow{1}{*}{ARM} 
    % & &  LSTM {\footnotesize \citep{Cooijmans2017recurrentbatch}} & 1.42 & - & 19.8$^\dagger$ \\
    % & &  Large mLSTM {\footnotesize \citep{krause2017multiplicativelstm}} & 1.27 & 1.24 & - \\
    & & 64 Layer Transformer {\footnotesize \citep{alrfou2019transformer}} & 35.5$^\dagger$ \\ \midrule
    % & & Sparse Transformer {\footnotesize \citep{child2019sparse}} & - & 0.99 & - \\ \cline{3-6}
   \multirow{2}{*}{\small VAE} & & AF/AF$^\star$ (AR) {\footnotesize \citep{ziegler2019latent}} & 156 {\small$\pm 1.8$} \\
   & & IAF / SCF$^\star$ {\footnotesize \citep{ziegler2019latent}} & 0.04 {\footnotesize $\pm0.004$} \\ \midrule
    \multirow{3}{*}{Generative Flow}
    & & Argmax Flow, AR (ours) & 115 {\small$\pm 0.03$} \\
    & & Argmax Coupling Flow (ours) & 0.40 {\footnotesize $\pm0.03$} \\
    % \multirow{4}{*}{\rotatebox[origin=c]{90}{\small \textit{Non-AR}}} & &  \\
    & & Discrete Flow {\footnotesize \citep{discrete2019tran}} & 0.16$^\dagger$\hspace{.75cm} \\ \midrule
    % & & Argmax Coupling Flow (ours) & 1.80 & 1.94 & 0.40 {\footnotesize $\pm0.03$} \\ 
    \multirow{2}{*}{Diffusion} & & Multinomial Text Diffusion (ours) & 26.6 {\small $\pm2.2$}$^{\ddagger}$ \\
     & & Multinomial Text Diffusion, 100 steps (ours) & 2.4 {\small $\pm0.16$} \\ \bottomrule \end{tabular}
\begin{flushleft}
 \scriptsize{$\dagger$ Computed on a 288-length sequence instead of 256-length, taken from \citep{discrete2019tran}.} \\
 \scriptsize{$\ddagger$ This result is for the complete 1000 timesteps chain, improvements are possible by skipping steps.}
 \vspace{-.5cm}
\end{flushleft}
\end{table}

Due to the computational cost of running normalizing flows, it is not possible for us to run every model many times. However, generally single-run results suffice, as the performance variance of these models is relatively small. In Table~\ref{tab:stdev} the standard deviation and average performance for a selection of models is shown, taken over $3$ runs. Observe that these standard deviations are small compared to the reported differences between the models. Notice that standard deviations for coupling models are larger, but the performance difference between those types of models is also larger.

\begin{table}[H]
    \centering
    \vspace{-.25cm}
    \caption{Average and standard deviations of several models.}
    \label{tab:stdev}
    \scalebox{.9}{
    \begin{tabular}{l c c c c c}
    \toprule
    Dequantization & Flow type & Dataset & average & stdev \\ \midrule
    % Uniform dequantization  & \multirow{3}{*}{Autoregressive} & 1.90 & 2.14 \\
    % Variational dequantization  &  & 1.43 & 1.44 \\
    Argmax Flow (ours) &  AR & text8 & 1.38 & $0.001$ \\
    Argmax Flow (ours) &  AR & enwik8 & 1.42 & $0.008$ \\
    Argmax Flow (ours) &  Coupling & text8 & 1.82 & $0.017$ \\
    Argmax Flow (ours) &  Coupling & enwik8 & 1.93 & $0.012$ \\\bottomrule
    \end{tabular}}
\vspace{-.3cm}
\end{table}

Finally, we also compare argmax flows to a situation where its density model exactly matches the density model in \citep{lippe2020categorical} on \texttt{text8}. In this experiment Argmax Flows (1.43 bpc) outperform CategoricalNF (1.45 bpc) in an equal setting.

\newpage
\section{Samples from the text models}
\begin{figure}[H]
\centering
\includegraphics[width=.99\textwidth]{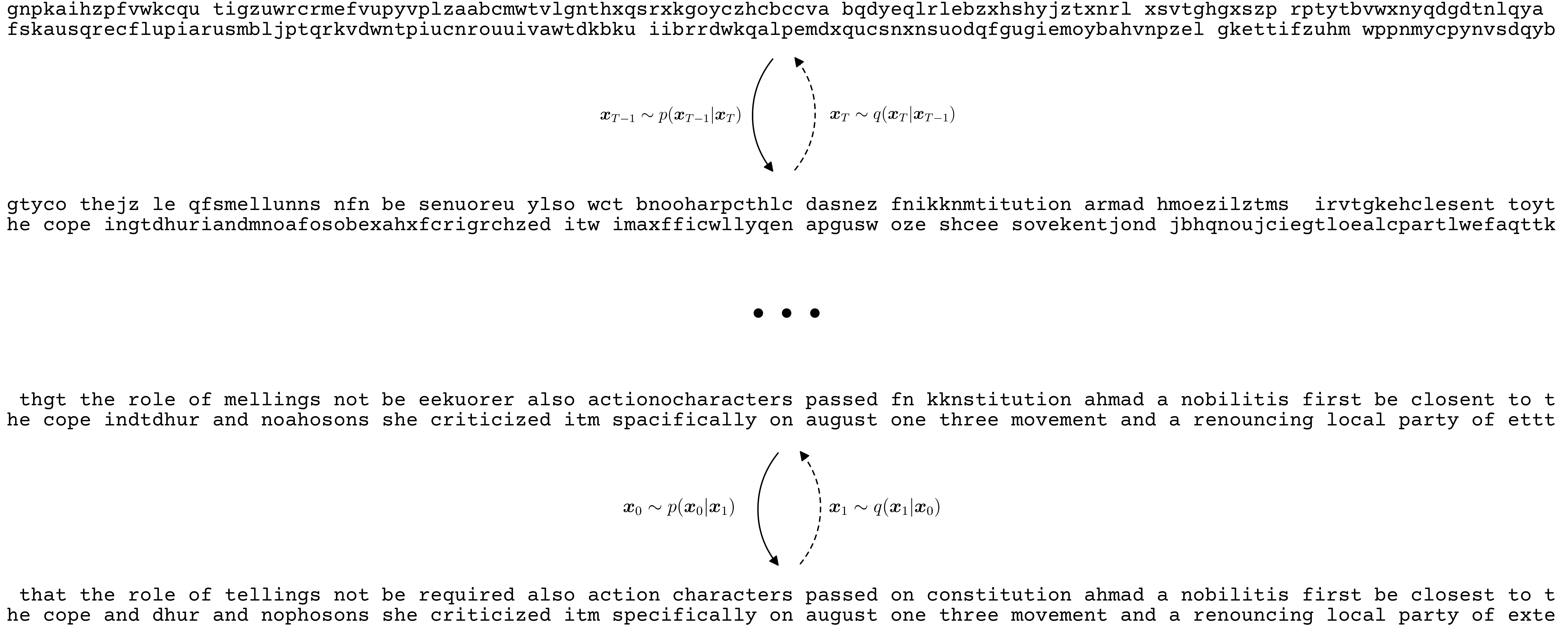}
\caption{Intermediate steps of the generation chain of the Multinomial Text Diffusion model trained on \texttt{text8}.}
\label{fig:samples_text_chain_appendix}
\end{figure}

Samples from our proposed models are presented in Table \ref{tab:samples_text_appendix} and a Multinomial Text Diffusion train is shown in Figure \ref{fig:samples_text_chain_appendix}, these results were not cherry-picked.

\begin{table}[H]
    \centering
    \caption{Samples from models trained on text8.}
    \label{tab:samples_text_appendix}
    \scalebox{.6}{
    \begin{tabular}{l l p{20cm}}
    \toprule
    Model & Nr & Text \\ \midrule
    \multirow{12}{*}{\rotatebox[origin=c]{90}{ \textit{Multinomial Diffusion}}} 
    & 1 & \texttt{ that the role of tellings not be required also action characters passed on constitution ahmad a nobilitis first be closest to the cope and dhur and nophosons she criticized itm specifically on august one three movement and a renouncing local party of exte} \\
    & 2 & \texttt{nt is in this meant the replicat today through the understanding element thinks the sometimes seven five his final form of contair you are lotur and me es to ultimately this work on the future all all machine the silon words thereis greatly usaged up not t} \\
    & 3 & \texttt{arity island louis has convinced privatist provinces the restrained marriage of his income ted guilds which in gulick performed in one nine six seven then sponly onward the bambat loving in separate including tichatta westell s doubled a bound of his futur} \\
    & 4 & \texttt{same early duration without education as a golden core power to the pirit of spain arriving wise speech art and r t plain firman q one five six the same as part of herald h rogenszers a art poetic of literature at shaft bressen three five three five eight } \\ \midrule
    \multirow{12}{*}{\rotatebox[origin=c]{90}{\textit{AR Argmax Flow}}} 
    & 1 & \texttt{ heartedness frege thematically infered by the famous existence of a function f from the laplace definition we can analyze a definition of binary operations with additional size so their functionality cannot be reviewed here there is no change because its } \\
    & 2 & \texttt{otal cost of learning objects from language to platonic linguistics examines why animate to indicate wild amphibious substances animal and marine life constituents of animals and bird sciences medieval biology biology and central medicine full discovery re} \\
    & 3 & \texttt{o use language combined with any of its subsets evolved into the group containing the primary concepts of a daily line on off the road and the material emulation of welcomes and prospects of pleasure and exercise have been committed projects in the economy} \\
    & 4 & \texttt{en that are beginning to forge since october one nine five zero the mandate was planted at k nigsberg during the car horizon at first please refer to a small government situated as well as in all these countries finally giving birth to a band here he was a} \\ \midrule
    \multirow{12}{*}{\rotatebox[origin=c]{90}{ \textit{Coupling Argmax Flow}}} 
    & 1 & \texttt{ns fergenur d alpha and le heigu man notabhe leglon lm n two six a gg
opa movement as sympathetic dutch the term bilirubhah acquired the bava
rian cheeh segt thmamouinaire vhvinus lihnos ineoneartis or medical iod
ine the rave wesp published harsy varb hhgh} \\
    & 2 & \texttt{and inequalities syllee mike jean  demet in standard rather than fmxed liga and a piare nut is gruncionde  aodadneveshiopyhabally uchc one viredtlty three ben yi agricultariis the only mefamantia or nuil and mid satio for kigou  wore not on the war rits af} \\
    & 3 & \texttt{e g chain within the sale of cooperative oppine p nge tyae yarot bouatta real frequency one mbj or rorbepetam iw by someone c langt b kindoms is the single yenta valve nor eosed collagen surkeys in the goubark cuisine of animum and two trantual measurement} \\
    & 4 & \texttt{hilepuin the king pete was added to or who cefralded to kiark n and panhpur not souhhvestern bat batas mudtlu for this creatures chew palenque lii lasron gentla tzanemi  derived from oo four issais nivissos with the name convertinus magaa named wes orieanr} \\ \midrule
    \end{tabular}}
\end{table}

\end{document}